\documentclass[sn-vancouver,Numbered]{sn-jnl}

\usepackage{graphicx}%
\usepackage{multirow}%
\usepackage{amsmath,amssymb,amsfonts}%
\usepackage{amsthm}%
\usepackage{mathrsfs}%
\usepackage[title]{appendix}%
\usepackage{xcolor}%
\usepackage{textcomp}%
\usepackage{manyfoot}%
\usepackage{booktabs}%
\usepackage{algorithm}%
\usepackage{algorithmicx}%
\usepackage{algpseudocode}%
\usepackage{listings}%
\usepackage{todonotes}

\usepackage{subcaption}

\usepackage[T1,T2A]{fontenc}
\usepackage[utf8]{inputenc}

\begin{document}

\title[Synthesizing Evolving Symbolic Representations for Autonomous Systems]{Synthesizing Evolving Symbolic Representations for Autonomous Systems}


\author*[1]{\fnm{Gabriele} \sur{Sartor}}\email{gabriele.sartor@unito.it}

\author[2]{\fnm{Angelo} \sur{Oddi}}\email{angelo.oddi@istc.cnr.it}

\author[2]{\fnm{Riccardo} \sur{Rasconi}}\email{riccardo.rasconi@istc.cnr.it}

\author[2]{\fnm{Vieri Giuliano} \sur{Santucci}}\email{vieri.santucci@istc.cnr.it}

\author[1]{\fnm{Rosa} \sur{Meo}}\email{rosa.meo@unito.it}

\affil*[1]{\orgdiv{Computer Science Department}, \orgname{University of Turin}, \orgaddress{\street{Via Verdi 8}, \city{Turin}, \postcode{10124}, \state{}, \country{Italy}}}

\affil[2]{\orgname{Institute for Cognitive Sciences and Technologies}, \orgaddress{\street{Via G. Romagnosi 18A}, \city{Rome}, \postcode{00196}, \state{}, \country{Italy}}}


\abstract{
In recent years, Artificial Intelligence (AI) systems have made remarkable progress in various tasks. Deep Reinforcement Learning (DRL) is an effective tool for agents to learn policies in low-level state spaces to solve highly complex tasks. Recently, researchers have introduced Intrinsic Motivation (IM) to the RL mechanism, which simulates the agent's curiosity, encouraging agents to explore interesting areas of the environment. This new feature has proved vital in enabling agents to learn policies without being given specific goals.

However, even though DRL intelligence emerges through a sub-symbolic model, there is still a need for a sort of abstraction to understand the knowledge collected by the agent. 
To this end, the classical planning formalism has been used in recent research to explicitly represent the knowledge an autonomous agent acquires and effectively reach extrinsic goals. 
Despite classical planning usually presents limited expressive capabilities, Probabilistic Planning Domain Definition Language (PPDDL) demonstrated usefulness in reviewing the knowledge gathered by an autonomous system, making explicit causal correlations, and can be exploited to find a plan to reach any state the agent faces during its experience.

This work presents a new architecture implementing an open-ended learning system able to synthesize from scratch its experience into a PPDDL representation and update it over time.
Without a predefined set of goals and tasks, the system integrates intrinsic motivations to explore the environment in a self-directed way, exploiting the high-level knowledge acquired during its experience.
The system explores the environment and iteratively: (a) discover options, (b) explore the environment using options, (c) abstract the knowledge collected and (d) plan.
This paper proposes an alternative approach to implementing open-ended learning architectures exploiting low-level and high-level representations to extend its own knowledge in a virtuous loop.
}




\maketitle

\section{Introduction}
    \label{sec1}
In the last few years, new AI systems have solved incredible tasks.
These tasks include real-world games, such as chess \cite{deepblue2002} and Go \cite{silver2018alphazero,silver2016mastering,silver2017mastering}, videogames such as Atari \cite{oh2015action}, Dota \cite{berner2019dota}, and different robotics tasks \cite{koch2019reinforcement,nguyen2019review,ibarz2021train,bellemare2020autonomous}.
These results have been mostly achieved through the intensive use of Reinforcement Learning (RL, \cite{sutton2018reinforcement}) with the rediscovered technology of neural networks and deep learning \cite{bengio2017deep}.
Usually, ``standard'' RL focuses on acquiring policies that maximise the achievement of fixed assigned tasks (through reward maximisation) with a predefined collection of skills.
New approaches have been proposed to enrich RL, allowing the agent to extend its initial capabilities over time inspired by neuroscience and psychology.
Indeed, studies on animals \cite{butler1953discrimination,harlow1950learning,montgomery1954role} and humans \cite{berlyne1950novelty,berlyne1966curiosity,RYAN2000motivantion} have explored the inherent inclination towards novelty, which is further supported by neuroscience experiments \cite{duzel2010novelty,ranganath2003neural,redgrave2006short}.
The field of intrinsically motivated open-ended learning (IMOL \cite{Santucci2020}) tackles the problem of developing agents that aim at improving their capabilities to interact with the environment without any specific assigned task. 
More precisely, Intrinsic Motivations (IMs \cite{Oudeyer2007intrinsic,Baldassarre2013Book}) are a class of self-generated signals that have been used to provide robots with autonomous guidance for several different processes, from state-and-action space exploration \cite{Frank2014,Bellemare2016}, to the autonomous discovery, selection and learning of multiple goals \cite{grail_2016,Colas2019,blaes2019control}. 
In general, IMs guide the agent in acquiring new knowledge independently (or even in the absence) of any assigned task to support open-ended learning processes \cite{Sigaud2023OEL}. 
This knowledge will then be available to the system to solve user-assigned tasks \cite{Baldassarre2024Purpose} or as a scaffolding to acquire new knowledge cumulatively \cite{SinghBartoChentanez2004,Forestier2017,Romero2022Curriculum} (similarly to what has been called curriculum learning \cite{Bengio2009}).

The option framework has been combined with IMs and ``curiosity-driven'' approaches to drive option learning \cite{SinghBartoChentanez2004} and option discovery \cite{Machado2017,oddi2020integrating,SartorZMORS21,sartor2022option}. 
In the hierarchical RL setting \cite{barto2003HRL}, where agents must chunk together different options to properly achieve complex tasks, IMs have been used to foster sub-task discovery and learning \cite{Konidaris2009,Niel2018,Rafati2019HRL}, and exploration \cite{Bellemare2016}. 
Autonomously learning and combining different skills is a crucial problem for agents acting in complex environments, where task solving consists of achieving several (possibly unknown) intermediate sub-tasks that are dependent on each other. 
An increasing number of works are tackling this problem \cite{blaes2019control,Parisi2021,Romero2021}, most focused on low-level, sub-symbolic policy learning \cite{bagaria21a}, in turn combined in a hierarchical manner using some sort of meta-policy \cite{veeriah2021discovery}. 
While promising, these approaches necessarily face the problem of exploration, which becomes slower and less efficient as the space of states and actions increases.

In contrast to sub-symbolic methods, symbolic approaches like Automated Planning \cite{NauGhallabTraverso2004,russell2010artificial} enable the use of higher-level objects (referred to as symbols), resulting in quicker execution, facilitating the composition of complex sub-task sequences, and making the agent's internal knowledge interpretable.
However, Automated Planning approaches require that the high-level representation of the planning domain is appropriately defined in advance.
Generally, planning requires prior knowledge of the world in which the agent operates expressed in terms of both the preconditions necessary for the execution of the actions as well as the effects that follow from executing them.
The need to be provided with an \textit{ad-hoc} symbolic representation of the environment limits the utilization of high-level planning for artificial agents in unknown or highly unstructured settings, where the acquisition of new knowledge and new skills is the progressive result of the agent's autonomous exploration of the environment.
However, some works tried to improve classical planning with autonomous model learning \cite{chitnis2021glib}, suggesting to add new symbols to the symbolic representation supported by the human \cite{mikita2012interactive} and using a cognitive layer to manage an intermediate representation \cite{rodriguez2018generating}.

Recently, some ideas have appeared in the literature proposing methodologies for integrating sub-symbolic and symbolic approaches, or more generally, low-level and high-level modules \cite{frossi2022combining}.
On the one hand, some works tried to reconcile deep learning with planning \cite{garnelo2016towards,GARNELO201917}, goal recognition \cite{amado2018goal} and the synthesis of a symbolic representation of the domain \cite{Asai_Fukunaga_2018,asai2019unsupervised}.
On the other hand, the integration has also been performed through a specific algorithm designed to produce an automated symbolic abstraction of the low-level information acquired by an exploring agent~\cite{konidaris2018skills} in terms of a high-level planning representation such as the PDDL formalism \cite{Ghallab98}, which explicitly describes  the context necessary to execute an action on the current state (i.e., the \textit{preconditions} and the \textit{effects}) making use of symbols.
This algorithm has been used as a module in architectures that integrate abstraction, planning and intrinsic motivations, such as IMPACT \cite{oddi2019learning,oddi2020integrating}.

This work presents an approach to the integration of low-level skills and high-level representations that allows to continuously update the set of low-level capabilities and their corresponding abstract representations.
Both the creation and the extension of the agent's knowledge is based on the implementation of two forms of intrinsic motivation (IM) which, respectively, (i) drive the agent to learn new policies while exploring the environment and (ii) encourage it to use its skills to reach less explored states, the rationale being that exploring unknown states increases the likelihood to learn new skills.
Then, the data collected by the agent's sensors before and after the execution of its skills are used by a specific algorithm to synthesize an updated abstract representation which can be used to plan the execution of sequences of low-level skills to reach more complex goals.
The main contribution of this study is to create a framework that, virtually starting from zero symbolic knowledge, produces an abstraction of the low-level data acquired from the agent's sensors, whose enhanced expressiveness can be exploited to plan sequences of actions that reach more and more complex goals.

\section{Background}
    \label{sec2}
To reach a high level of autonomy, an agent acting in the low-level space, sensing the environment with its sensors and modifying it through its actuators must implement a series of layers of abstraction over its state and action spaces.
As human beings reason over both simple and complex concepts to perform their activities, so robots should be able to build their own abstract representation of the world to deal with the increased complexity, using \textit{labels} to refer to actions and events to be recognized and reasoned upon.
In this paper, two level of abstractions are applied: the first one, from primitive actions to \textit{options} \cite{SUTTON1999181_framework_b,sutton2018reinforcement} and the second one, from options to classical planning \cite{russell2010artificial}.

\subsection{From primitives to options}
\label{subsec:options}
As discussed before, at the lowest level, the agent sees the world with its sensor's values and changes it through the movement of its actuators.
The most common formalism at this stage to deal with this type of representation is the Markov Decision Process (MDP), which models the environment as the tuple:
\begin{equation}
    (S,A,R,T,\gamma), \label{eq:mdp}
\end{equation}
in which $S$ represents the set of possible high-dimensional states where each $s \in S$ is described by a vector of real values returned by the agent's sensors, $A$ describes the set of low-level actions $a \in A$ in some cases also called \textit{primitives}, $R$ the reward function where $R(s,a,s')$ is a real value returned executing $a$ from state $s$ achieving $s'$, $T$ the transition function describing for $T(s'|s,a)$ the probability of reaching the state $s'$ executing $a$ from $s$, and the discount factor $\gamma \in (0,1]$ describing the agent's preference for immediate over future rewards.
Usually, in this setting, the goal is to maximize \textit{return}, defined as
\begin{equation}
    \mathcal{R} = \sum_{i=0}^{\infty} \gamma^i R (s_i,a_i,s_{i+1}).
\end{equation}

However, dealing with the state and action spaces of the formulation (\ref{eq:mdp}) is, in certain cases, impractical due to the high dimensional spaces considered.
An effective formalism introduced to reduce the complexity of the problems is the \textit{option} framework \cite{SUTTON1999181_framework_b}.
The \textit{option} is a temporally-extended action definition which employs the following abstracted representation:
\begin{equation}
    o = (I_o, \pi_o, \beta_o),
\end{equation}
where the option $o$ is defined by an \textit{initiation set} $I_o = \{s | o \in O(s)\}$ representing the set of states in which $o$ can be executed, a \textit{termination condition} $\beta_o(s) \rightarrow [0,1]$ returning the probability of termination upon reaching $s$ by $o$, and a policy $\pi_o$ which can be run from a state $s \in I_o$ and terminated reaching $s'$ such that the probability $\beta_o(s')$ is sufficiently high.
A \textit{policy} is a function defining the behavior of the agent, mapping the perceived state of the environment to the action to be taken \cite{sutton2018reinforcement}.
It is worth noting that options create a temporally-extended definition of the actions \cite{SUTTON1999181_framework_b}.
Indeed, the option is an abstraction defining an action as a repeated execution of policy $\pi$ from a state $s \in I_o$ to another $s'$ in a maximum amount of time steps $\tau$.

\begin{figure}
    \centering
    \begin{subfigure}[b]{0.3\textwidth}
        \centering
        \includegraphics[width=\textwidth]{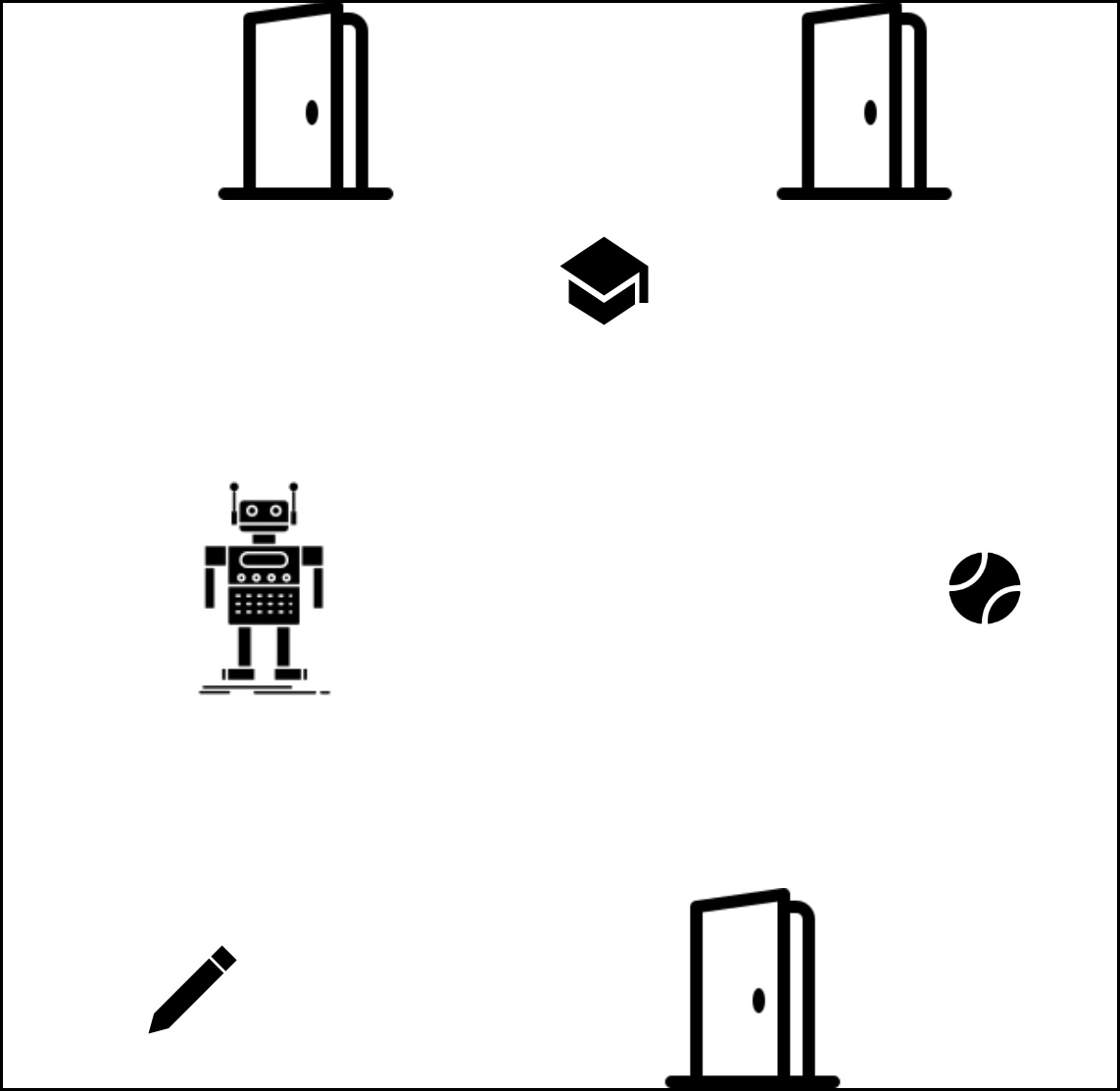}
        \caption{}
            \label{option_example_a}
    \end{subfigure}
    \begin{subfigure}[b]{0.3\textwidth}
        \centering
        \includegraphics[width=\textwidth]{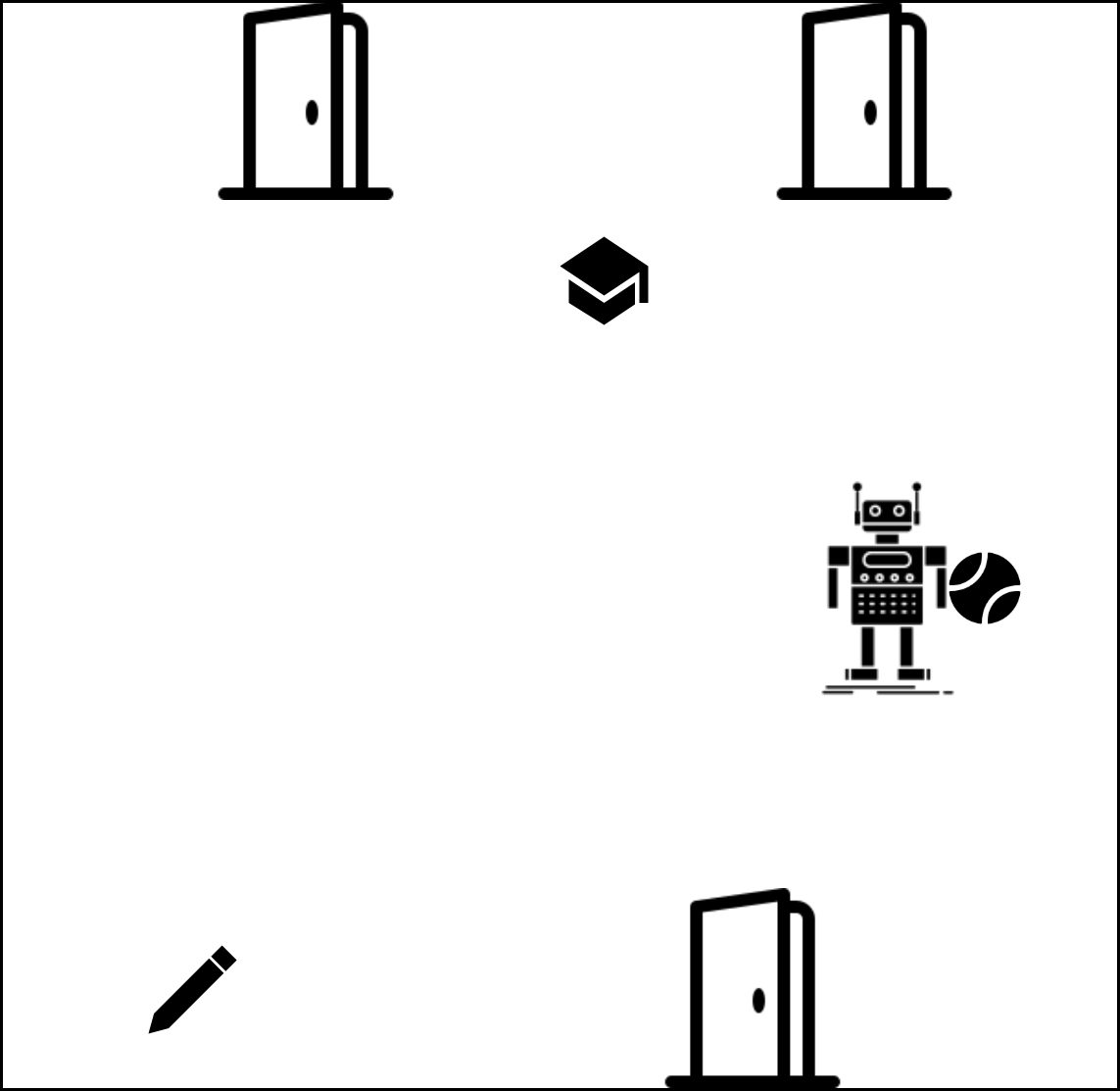}
        \caption{}
        \label{option_example_b}
    \end{subfigure}
       \caption[Representation of the option \textit{reach the ball}.]{Representation of the option \textit{reach the ball}.}
       \label{fig:option_example}
\end{figure}

For instance, Figure \ref{fig:option_example} depicts an environment before and after executing the option \textit{``reach the ball''}, respectively Figure  \ref{option_example_a} and  \ref{option_example_b}.
When the agent wants to execute this option, it checks whether its current state belongs to the \textit{initiation set} of the option.
In our case, and assuming that the option's policy simply consists in moving towards the ball, the agent can execute the option and runs the policy $\pi_o$ until the \textit{termination condition} returns a sufficiently high probability of success or $\tau$ time steps are reached.
In the Figure \ref{option_example_b} the options successfully terminates getting close to the ball.

Passing from low-level actions to options reduces the agent's action space.
Adopting options in the MDP formalism implies moving to the semi-Markov Decision Process (SMDP):
\begin{equation}
    (S,O,R,P,\gamma),
\end{equation}
where $S$ is the original state space, $O(s)$ is the set of options executable from state $s$, $R(s,\tau | s,o)$ describes the reward expected executing $o \in O(s)$ from state $s$ reaching $s'$ after $\tau$ time steps, $P(s',\tau | s,o)$ returns the probability of reaching state $s'$ after $\tau$ time steps executing $o \in O(s)$ from state $s$, and the discount factor $\gamma \in (0,1]$.
Using options entails moving in the low-level state space $S$ with abstracted actions permitting the agent to perform extended and more complex behaviors, reducing the number of actions to achieve a certain task and simplifying the problem.

\subsection{Options and Classical Planning}\label{subsec:options_in_classical_planning}
The option formalism and its way of abstracting the dynamics of an environment share common characteristics with \textit{classical planning}, in which the world is described in a simplified formal description considering only the aspects necessary to solve the agent's task \cite{russell2010artificial}.
Planning is the field of research studying formal methods to automatically find solutions, also called plans, to tasks requiring a sequence of actions $[\alpha_1,\dots,\alpha_n]$ to reach a goal state $s_g$ from an initial state $s_{init}$.
The plan $\omega$ is obtained by giving in input a model of the environment dynamics and the problem definition to a planner, returning a solution applying general optimization algorithms.

\textit{Classical planning} is a particular instance of this methodology exploiting a logical language (i.e. propositional logic, first-order logic, etc.) to capture an abstract symbolic description of the world \cite{russell2010artificial}.
The \textit{symbol}, which is the core of classical planning, is a name given to a certain set of states $s \in S$ satisfying a specific condition in the sensorimotor space.
This \textit{mapping} from the symbol to the real world states is called \textit{grounding}.
Specifically, a symbol $\sigma_Z \in \Sigma$, with $\Sigma$ set of the available symbols, is the name given to a \textit{test} $\tau_Z$, and the corresponding set of low-level states where the test is satisfied $Z = \{s \in S \mid \tau_Z(s) = True\}$, with the high-dimensional low-level state space $S$ \cite{konidaris2018skills}.
To determine whether or not a low-level state $s_i$ belongs to the state set $Z$, the $\sigma_Z(s_i)$ test is run, which will return either $True$ or $False$.
Again, both $Z$ and $\sigma_Z$ provide the \textit{semantics} of the symbol; $Z$ is the symbol's \textit{grounding set}, while $\sigma_Z$ is its \textit{grounding classifier}.
Symbols can be combined using \textit{operators} having the following meaning:
\begin{itemize}
	\item $\neg \sigma_Z$ corresponds to the \textit{negation} of symbol $\sigma_Z$;
	\item $\sigma_X \vee \sigma_Y$ corresponds to the \textit{union} of symbols $\sigma_X$ and $\sigma_Y$ (i.e.,  the union of their respective \textit{grounding sets});
	\item $\sigma_X \wedge \sigma_Y$ corresponds to the \textit{intersection} of symbols $\sigma_X$ and $\sigma_Y$ (i.e., of their respective \textit{grounding sets}).
\end{itemize}
In classical planning, operators and symbols are used to describe actions in the following form
\begin{equation}
    \alpha_i = (pre_i,eff_i^+,eff_i^-), \label{eq:action} 
\end{equation}
meaning that the action $\alpha_i \in \mathcal{A}$ can be executed when all the symbols $\{\sigma | \sigma \in pre_i \}$, also called \textit{preconditions}, are \textit{True}, and executing $\alpha_i$ produces the changes of the value of some symbols, also called \textit{effects}, implying that all symbols $\{\sigma | \sigma \in eff_i^+ \}$ assume the value \textit{True} and symbols $\{\sigma | \sigma \in eff_i^- \}$ become \textit{False}.

Finally, using symbols $\Sigma$ and high-level actions $\mathcal{A}$ as building blocks, it is possible to define the model of environment $\mathcal{D}$, also called \textit{domain}, and the \textit{problem} $\mathcal{P}$ to solve.
A classical planning domain can be defined as
\begin{equation}
    \mathcal{D} = (\Sigma, \mathcal{A}, \Gamma),
\end{equation}
using a set of symbols $\Sigma$, actions $\mathcal{A}$ and a \textit{state-transition function} $\Gamma : \hat{\Sigma} \times \mathcal{A} \rightarrow \hat{\Sigma}$, where $\hat{\Sigma}$ is the set of possible subsets of $\Sigma$.
The \textit{state-transition} function $\Gamma(\hat{\Sigma}_{s},\alpha_i) = (\hat{\Sigma}_{s} - eff_i^-) \cup eff_i^+$, if $\alpha_i$ is applicable to $\hat{\Sigma}_{s}$, where $\hat{\Sigma}_{s}$ is the set of symbols whose \textit{grounding set} intersection defines the state $s \in S$.
These elements are sufficient to describe the dynamics of the environment, over which the planner can reason and create chains of actions to reach a final goal.
The function $\Gamma$ encapsulates the transition model of the environment in each action model described by (\ref{eq:action}), defining the possible way to build sequences of them.
Instead, the problem can be formalized as 
\begin{equation}
    \mathcal{P} = (\hat{\Sigma}_{s_{init}}, \hat{\Sigma}_{s_{g}} ),
\end{equation}
where $\hat{\Sigma}_{s_{init}}$ is the set of symbols whose \textit{grounding set} intersection describes $s_{init}$ 
and $\hat{\Sigma}_{s_{g}}$ the set of symbols whose \textit{grounding set} intersection characterizes $s_{g}$.
Then, the plan solving the problem $\mathcal{P}$ is expressed as the sequence of actions
\begin{equation}
    \omega = [\alpha_1,...,\alpha_n],
\end{equation}
resulting in a sequence of state transitions
\begin{equation}
    [\hat{\Sigma}_{s_{0}},\hat{\Sigma}_{s_1},\dots,\hat{\Sigma}_{s_{n}}],
\end{equation}
such that $\Gamma(\hat{\Sigma}_{s_{0}},\alpha_1) = \hat{\Sigma}_{s_{1}}, \Gamma(\hat{\Sigma}_{s_{1}},\alpha_2) = \hat{\Sigma}_{s_{2}}, \dots, \Gamma(\hat{\Sigma}_{s_{n-1}},\alpha_n) = \hat{\Sigma}_{s_{n}}$ with initial state $\hat{\Sigma}_{s_0} = \hat{\Sigma}_{s_{init}}$ and final state $\hat{\Sigma}_{s_n} = \hat{\Sigma}_{s_g}$.
In particular, the formulation presented based on a finite set of symbols $\Sigma$ and state-transition system $\mathcal{T} = (\Sigma,\mathcal{A},\Gamma)$ is called a \textit{set-theoretic representation} \cite{russell2010artificial}.
In order to use classical planning systems, it is necessary to describe both the domain of the considered world and the tackled problem using a planning definition language.
In this work we will use the Probabilistic Planning Domain Definition Language (PPDDL) \cite{younes2004ppddl1}; once defined, both the domain and the problem definitions will be provided in input to the planner, which will eventually return a solution $\omega$.

It is worth noting that the option $o$ and the planning action $\alpha$ share some similarities.
Indeed, $o$ can be converted in its corresponding high-level action $\alpha$ finding the right set of symbols $pre,eff^+,eff^-$ whose grounding sets satisfy the classifiers $I_o$ as preconditions and $\beta_o$ as effects, for the execution of $\pi_o$.
The similarity between options and set-theoretic planning actions has been exploited to create automatic abstraction procedures able to convert the options' execution data into a working high-level planning description, thus allowing a solution that integrates ating low-level and high-level information.
In this work we build upon the abstraction procedure implemented by Konidaris et al. \cite{konidaris2018skills} to convert sensors raw data into a PPDDL representation.

\subsection{Intrinsic Motivation}
The impulse to drive the agent away from the monotony of its usual activities, which psychologists and cognitive scientists have studied under the name of \textit{intrinsic motivation}, is one of the most important elements enabling Open Ended Learning (OEL). 
The research in the field of Intrinsic Motivation (IM) concerns the study of human behaviors not influenced by external factors but characterized by internal stimuli (i.e. curiosity, exploration, novelty, surprise).
In the case of artificial agents, we can summarize such aspect as anything that can drive the agent's behavior which is not directly dependent on its assigned task.

The insights provided by the IMs gave the researchers new ideas to model the stimuli of the agent (e.g. curiosity).
Indeed, some models have been implemented using the prediction error (PE) in anticipating the effect of agent's actions (and more precisely the improvement of the prediction error \cite{schmidhuber2010formal,Santucci2010biological}) as an IM signal.
A formal definition of agent driven by its curiosity has been formulated by Schmidhuber \cite{schmidhuber2010formal} as simply maximizing its \textit{future success} or \textit{utility}, which is the conditional expectation
\begin{equation}
    u(t) = E_\mu \left[ \sum_{\tau = t+1}^{T} r(\tau) \bigg| h(\leq t) \right], \label{eq:intrinsic_function}
\end{equation}
over $t = 1,2,\dots,T$ time steps, receiving in input a vector $x(t)$, executing the action $y(t)$,
returning the reward $r(t)$ at time $t$, taking into consideration the triplets $h(t) = [x(t),y(t),r(t)]$ as the previous data experienced until time step $t$ (also called \textit{history}).
The conditional expection $E_\mu(\cdot|\cdot)$ assume an unknown probability distribution $\mu$ from $M$ representing possible probabilistic reactions of the environment.
To maximize (\ref{eq:intrinsic_function}), the agent also has to build a predictor $p(t)$ of the environment to anticipate the effects of its actions.
The reward signal is defined as follows
\begin{equation}
    r(t) = g(r_{ext}(t),r_{int}(t)),
\end{equation}
which is a certain combination $g$ of an external reward $r_{ext}(t)$ and an intrinsic reward $r_{int}(t)$.
In particular, $r_{int}(t+1)$ is seen as \textit{surprise} or \textit{novelty} in assessing the improvements in the results of $p$ at time $t+1$
\begin{equation}
    r_{int}(t+1) = f | C(p(t),h(\leq t+1)), C(p(t+1),h(\leq t+1)) |,
\end{equation}
where $C(p,h)$ is a function evaluating the \textit{performance} of $p$ on a history $h$ and $f$ is a function combining its two parameters (e.g. in this case, it could be simply the improvement $f(a,b) = a - b$).
It is important to notice that, as a baby does, an intrinsically motivated agent needs to find regularities in the environment to learn something.
Consequently, if something does not present a \textit{pattern}, there is nothing to learn and this becomes boring for both an agent and a human being.

In literature, IMs have also been categorized into different typologies \cite{barto2013novelty,mirolli2013functions,oudeyer2009intrinsic}.
An important discriminant aspect is the kind of signal received by the agent which can be of two types: \textit{knowledge-based} (KB-IMs), which depends on the prediction model of the world (e.g. \cite{schmidhuber2010formal}), and \textit{competence-based} (CB-IMs), which depends on the improvement of the agent's skills (e.g. \cite{grail_2016}).
In the framework presented in the next section, both these typologies are employed. 
CB-IM is used at a lower level to learn new skills and KB-IM at higher level to push the system to focus on the frontier of the visited states, from which it is more likely to discover novel information (e.g., find new states and learn new actions).

\section{System Overview}
\label{sec3}
This section presents a new framework of an open-ended learning agent which, starting from a set of action primitives, is able to (i) discover options, (ii) explore the environment using them, (iii) create a PPDDL representation of the collected knowledge and (iv) plan to improve its exploration while reaching a high-level objective set by the game.

The aim of this study is to assess the potential of \textit{abstraction} in autonomous systems and propose a new approach for planning systems, extending them with learning capabilities and behaviors driven by IMs.
IMs are employed for discovering new options in a \textit{surprise-based} manner at low-level and continuously exploring new states driven by \textit{curiosity} at high-level.
By the term abstraction, we simply mean mapping a certain problem space into a simpler one (e.g. converting a continuous domain into a discrete domain).
In the proposed architecture, the abstraction is applied at two levels: a) passing from the \textit{primitive action space} to the \textit{options action space} and b) converting \textit{low-level data} collected during the exploration into a \textit{high-level domain representation} suitable for high-level planning, thus from raw sensors' data to a PPDDL representation.

\begin{figure}[ht]
    \centering
    \includegraphics[width=0.98\textwidth]{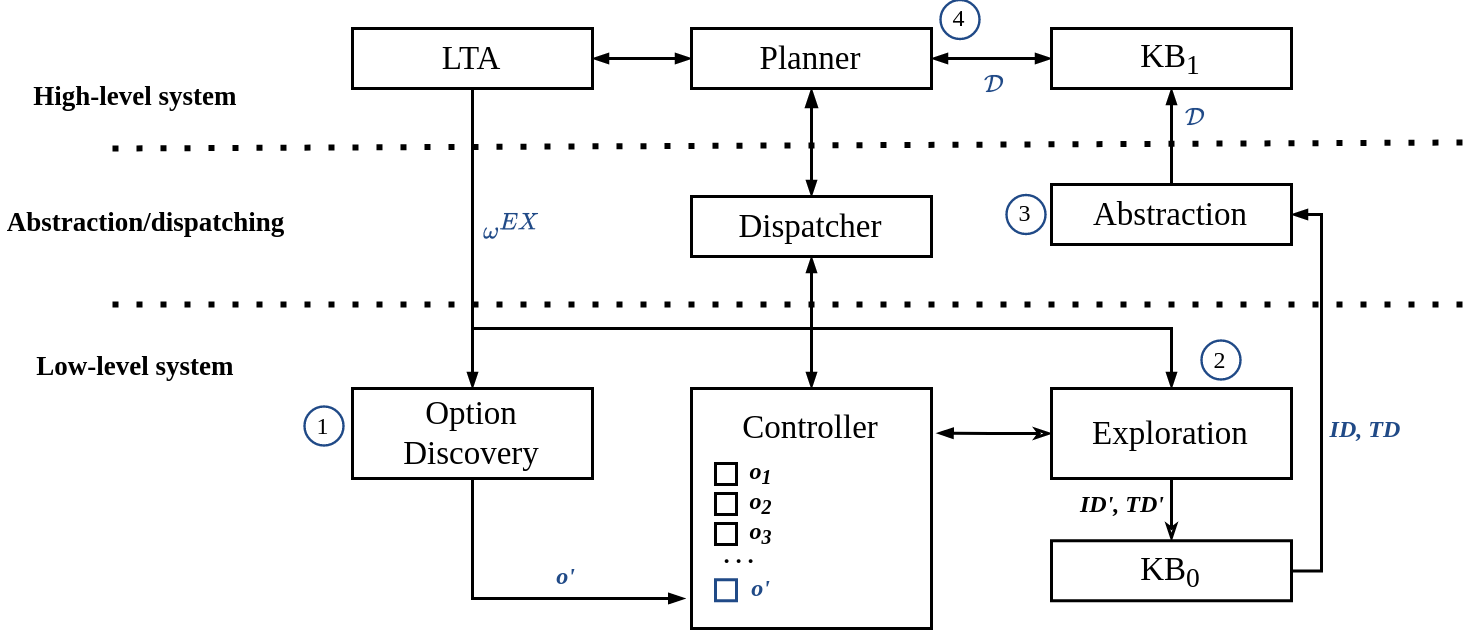}
    \caption[The architecture of the intrinsically motivated agent using high-level planning for its exploration.]{The architecture of the system. The system, equipped with five primitive actions, iteratively (1) learns some options from scratch, (2) uses the options to explore the environment and gather low-level states data, (3) creates a PPDDL representation of the collected experience, (4) plan to solve the game and to improve its exploration.}
    \label{fig:dpa_architecture}
\end{figure}

Performing the pipeline depicted in Figure \ref{fig:dpa_architecture}, the system creates different layers of abstraction, enriching the agent's knowledge with causal correlations between options and enabling more efficient reasoning (i.e. using classical planning).
Symbols can be seen as knowledge building blocks that can be used to search for interesting states and find new knowledge in a virtuous loop.

\subsection{Architecture description}
As depicted in Figure \ref{fig:dpa_architecture}, the system can be seen as a three-layered architecture: (i) the higher level contains the explicit agent's knowledge, (ii) the middle layer maps the high-level actions to their controllers and convert the raw sensors data into an explicit representation, and (iii) the lower level containing the components to sense the environment and interact with it.
Mainly, the system executes the following pipeline:
\begin{enumerate}
    \item \textbf{Option Discovery:} using a set of primitives $A = \{a_1,a_2,...\}$ belonging to the agent, the system combines them to create higher level actions, in this case, the \textit{options};
    \item \textbf{Exploration:} the set of options $O = \{o_1,o_2,...\}$ discovered in the previous step is used to explore the environment. In the meantime, the visited low-level states data are collected into two datasets: the \textit{initiation data} $ID$ and \textit{transition data} $TD$ containing, respectively, the samples of states in which an option can be run and the transition between states executing it;
    \item \textbf{Abstraction:} datasets $ID$ and $TD$ of the visited low-level states until that moment are processed by the algorithm of abstraction, generating a PPDDL domain $\mathcal{D}$ of the knowledge collected;
    \item \textbf{Planning:} the PPDDL representation $\mathcal{D}$ is used to assess whether the final goal of the task $s_g$ can be reached with the currently synthesized knowledge, and to generate a plan $\omega^{EX}$ to explore interesting areas of the domain. This plan is suggested by the \textbf{Goal Selector}, function included in the Long-Term Autonomy (LTA) module.
    \item The system, coordinated by the LTA, will execute again the loop from step 1, exploiting $\omega^{EX}$ to improve the Option Discovery and Exploration phases.
\end{enumerate}

\begin{algorithm}
\scriptsize
\caption{Discover-Plan-Act algorithm}
\begin{algorithmic}[1]

\Procedure{DISCOVER\_PLAN\_ACT}{$cycles, dpa\_eps, dpa\_steps, d\_eps, d\_steps$}       
    \State $c \leftarrow 0$ //Cycle initialization
    \State $O \leftarrow \{\}$ //Option set initialization
    \State $ID \leftarrow \{\}$ //Initiation Data initialization
    \State $TD \leftarrow \{\}$ //Transition Data initialization
    \State $\omega^{EX} \leftarrow \{\}$ //Initially, the high-level plan is empty
    
    \While{$c < cycles$} //For each cycle \label{lst:framework_while}
        \State $O\_{new} \leftarrow DISCOVER(d\_eps, d\_steps, \omega^{EX}) $ \label{lst:discover} //Learning the available options
        \State $O \leftarrow O \cup O\_{new}$ \label{lst:option_union}
        \State $ID_{new}, TD_{new} \leftarrow Collect\_Data(dpa\_eps, dpa\_steps, O, \omega^{EX}) $ \label{lst:collect_data}
        \State $ID \leftarrow ID \cup ID_{new}$ \label{lst:init_data}
        \State $TD \leftarrow TD \cup TD_{new}$ \label{lst:transition_data}        
        \State $\mathcal{D} \leftarrow Create\_PPDDL(ID, TD)$ \label{lst:create_pddl}
        
        \State $s_{target} \leftarrow Get\_Target\_State()$\label{lst:get_target_state}
        
        \State $\mathcal{P}_{target} \leftarrow Generate\_PPDDL\_Problem(s_{target})$\label{lst:pddl_problem}
        
        \State $\omega^{EX} \leftarrow Plan(\mathcal{D}, \mathcal{P}_{target})$\label{lst:planning}

        \State $Check\_PPDDL\_Validity(\mathcal{D})$\label{lst:pddl_validity}
        
        \State $c \leftarrow c+1$ 
    \EndWhile \label{lst:framework_while_end}

\EndProcedure
\end{algorithmic} \label{alg:dpa_alg}
\end{algorithm}

In this setting, the agent is initially only endowed with a set of primitive movements $A = \{a_0,...,a_m\}$, and the world $s \in S$ is represented in terms of a vector $(v_0,...,v_n)$ of low-level variables $v_i \in \mathcal{R}$, whose values as retrieved by the agent's sensors.

The iterative utilization of this framework allows the synthesis of an emerging abstract representation of the world from the raw data collected by the agent, which continuously undergoes a refinement process over time, as it gets enriched with new actions and symbolic concepts.
In the following subsections, all the process funcionalities will be individually explained, with reference to the pseudocode depicted in Algorithm~\ref{alg:dpa_alg}.

\subsubsection{Option Discovery}\label{subsec:option_discovery}
In this section, we will analyze the \textbf{Option Discovery} module (Algorithm \ref{alg:dpa_alg}, line \ref{lst:discover}) in greater detail.
As previously anticipated, the discovery of new options is considered to be driven by the agent's \textit{surprise} in finding out that new primitives are available for execution, during the agent's operations.
When the agent encounters a change in the availability of its primitive abilities, it stores this event as a low-level skill that can be re-used later to explore the surrounding environment.

By executing the algorithm, the agent can discover a set of options $O$ from scratch; such options are generated by repeatedly executing a certain primitive $a \in A$ among the available ones and collecting the produced changes in the environment.
This procedure is intentionally implemented in a simplified way, given that the focus of this work is on the architecture for extensible symbolic knowledge to be reusable to reach intrinsic and extrinsic goals autonomously; more sophisticated stategies to discover new policies are left to future works.

In more detail, the agent creates new options considering the following modified definition of option:
\begin{equation}
    o(a^p, a^t, I, \pi, \beta),\label{eq:option}
\end{equation}
where $a^p$ and $a^t$ are primitive actions such that: (i) $a^p \neq a^t$, $a^p$ is used by the execution of $\pi$, (ii) $a^t$ stops the execution of $\pi$ when it becomes available, (iii) $\pi$ is the policy applied by the option, consisting in repeatedly executing $a^p$ until it can no longer be executed or $a^t$ becomes available, (iv) $I$ is the set of states from which $a^p$ can run; and (v) $\beta$ is the termination condition of the option, corresponding to the availability of the primitive $a^t$ or to the impossibility of further executing $a^p$.
For the sake of simplicity, in the remainder of the paper the option's definition will follow the more compact syntax 
\begin{equation}
    o(a^p,a^t)\label{eq:option_simplified}
\end{equation}
meaning that $I$ is the set of states in which $a^p$ can run, $\beta$ checks the following two conditions: $a^t$ becomes available or $a^p$ is no longer available, and $\pi$ is the policy corresponding to repeatedly executing $a^p$ until $\beta$ verifies.

Algorithm \ref{alg:discovery_alg} describes in further details the option discovery procedure previously described.
At the beginning of each discovery episode, the plan $\omega^{EX}$ is executed to reach a new area where to start learning new options (line \ref{lst:plan_im_execution_1}-\ref{lst:plan_im_execution_3}).
Then, for a maximum number of episodes and steps, the agent saves the current state $s$ and randomly selects a primitive $a^p$ which can be executed in $s$ (line \ref{lst:while}-\ref{lst:get_prim}). 
$a^p$ is repeatedly executed towards reaching the state $s'$ until either $a^p$ is no longer available or new primitives beyond $a^p$ become available.
If $s \neq s'$, the procedure creates a new option $o$ where $o = o(a^p,a^t)$ if a new primitive $a^t$ has become available, or $o = o(a^p,\{\})$ in the opposite case.
In either way, in case $o$ has not been discovered before, it is added to the other collected options.
It is important to note that the options are independent on the state where the agent is, and are defined by the primitives' availability.
This definition makes options reusable on different floors and with different objects, just depending on the agent's abilities.
The procedure can discover a subset of the options available in the original implementation\footnote{Link to the original implementation of \cite{konidaris2018skills}:  https://github.com/sd-james/skills-to-symbols} sufficient to solve the entire game problem.

\begin{algorithm}
\scriptsize
\caption{Option Discovery}
\begin{algorithmic}[1]

\Procedure{OPTION\_DISCOVERY}{$d\_eps, d\_steps, \omega^{EX}$}       
    \State $O_{new} \leftarrow \{\}$ 
    \State $ep \leftarrow 0$          

    \While{$ep < d\_eps$} //For each episode
        \State $T \leftarrow 0$
        \State $\textit{Reset\_Game()}$ 

        \For{$option \, \textbf{in} \, \omega^{EX}$} // \textit{Execute IM plan} \label{lst:plan_im_execution_1}
            \State $Execute(option)$            \label{lst:plan_im_execution_2}
        \EndFor                                 \label{lst:plan_im_execution_3}
        
        \While{$T < d\_steps$} \label{lst:while} //For each step
        \State $s \leftarrow \textit{Get\_State()}$ 
        \State $a^p \leftarrow \textit{Get\_Available\_Primitive()}$ \label{lst:get_prim}
        
            \While{$Is\_Available(a^p) \, \textbf{and not}\, (\textit{New\_Available\_Prim())}$} \label{lst:while_execution}
            \State $Execute(a^p)$
            \State $s' \leftarrow \textit{Get\_State()}$
            \EndWhile
            
            \If{$s \neq s'$}
                \If{$ \textit{New\_Available\_Prim()}$}
                    \State $a^t \leftarrow \textit{Get\_New\_Available\_Prim()}$
                    \State $o \leftarrow Create\_New\_Option(a^p, a^t)$ \label{lst:option_p_t}
                \Else
                    \State $o \leftarrow Create\_New\_Option(a^p, \{\})$ \label{lst:option_p_1}
                \EndIf
                \State $O_{new} \leftarrow O_{new} \cup o$ \label{lst:add_option}
            \EndIf
            \State $T \leftarrow T+1$ //End \textit{For each step}
        \EndWhile
        \State $ep \leftarrow ep+1$ //End \textit{For each episode}
    \EndWhile
    \State \Return $O_{new}$
\EndProcedure
\end{algorithmic} \label{alg:discovery_alg}
\end{algorithm}

\subsubsection{Exploration}
After discovering a set of valid options $O$ as explained in the previous section, the system exploits them to explore the environment, collecting data about the reached low-level states (Algorithm \ref{alg:dpa_alg}, line \ref{lst:collect_data}).
Considering that the abstracted representation of the world does not change significantly with a small amount of new data, the function \textit{Collect\_Data()} is in charge of executing \textit{d\_steps} options for \textit{d\_eps} episodes, in which the agent starts its exploration from the initial configuration of the environment.

At each timestep, the agent attempts to perform an option $o \in O$ from a certain low-level state $s \in S$.
The selection of the action $o$ during the exploration can follow different strategies, which are described in the subsection \ref{sub:intrinsic_motivation}.
In case the execution of the option changes the low-level variables of the state $s$ and, consequently, the \textit{mask}\footnote{The \textit{mask} is the list of all the state variables that are changed by the execution of a specific option. See details in  \cite{konidaris2018skills}.} \textit{m} is not null, the system registers two types of data tuple (Algorithm \ref{alg:dpa_alg}, line \ref{lst:collect_data}): the \textit{initiation data} tuple $id$ and the \textit{transition data} tuple $td$.
The multiple instances of these tuples are stored, respectively, in the datasets $ID$, for the initiation data, and $TD$, for transition data.
A single \textit{initiation data} tuple $id_i$ has the following structure 
\begin{equation}
    id_i = (s,o,f(s,o)),\label{eq:initiation}
\end{equation}
where the function $f(s,o)$ returns the feasibility of executing $o$ from $s$ ($True$ if $s \in I_{o}$ and $False$ otherwise).
The \textit{transition data} tuple $td_j$ takes the following structure
\begin{equation}
td_i = (s,o,r,s',g,m,O'),\label{eq:transition}
\end{equation}
where $s'$ is the state reached after executing option $o$ from the state $s$, $g$ is a flag stating whether the final objective of the task has been reached, $m$ is the \textit{mask} of the option and $O'$ is a list defining the options that can be executed from $s'$.
When all the steps of the episode have been executed, the environment is reset and the next episode is started until reaching the maximum number of allowed episodes \textit{d\_eps}, where the \textit{Collect\_Data()} procedure terminates and the stored data are added to the existing datasets $ID$ and $TD$ (line \ref{lst:init_data}-\ref{lst:transition_data}).

\subsubsection{Abstraction} \label{subsec:abstraction}
The datasets collected in the previous step are then used as input for the function $Create\_PPDDL()$ (Algorithm \ref{alg:dpa_alg}, line \ref{lst:create_pddl}), returning a symbolic representation $\mathcal{D}$ of the agent's current knowledge expressed in PPDDL formalism (PPDDL domain).
The main advantage of the obtained PPDDL representation is that it makes explicit the causal correlations between operators that would have remained implicit at the option level.
In the following, we provide a summary description of the abstraction procedure; for further details, the reader is referred to the original article~\cite{konidaris2018skills}.

The abstraction procedure executes the following five steps:
\begin{enumerate}
    \item \textbf{Options partition}: this step is dedicated to partitioning the learned options into \textit{abstract subgoal options}\footnote{An abstract subgoal option $o$ is characterized by a list of indices of the low-level variables, called \textit{mask}, which are changed with the execution of $o$, without modifying other variables. In addition, the changing variables' values do not depend on their initial value.}, a necessary assumption of the abstraction procedure. Abstract subgoal options are characterized by a single precondition and effect set.
    However, given the uncertainty of the actions' effects in the environment, the operators' effects will be modelled as \textit{mixture distributions} over states.
    This phase utilizes the transition dataset $TD$ collected before, as it captures the information about the domain segment the option modifies.
    Basically, the transition dataset is divided into sets of transition tuples presenting the same option $o$ and mask $m$.
    Then, for each set, the partitions are ultimately obtained by performing clustering on the final states $s'$ through the DBSCAN algorithm \cite{10.5555/3001460.3001507}.
    If some clusters overlap in their initiation set, a unique partition is created with different effects and occurrence probabilities.
    \item \textbf{Precondition estimation}: this step is dedicated to learning the classifiers that will identify the \textit{preconditions} of the PPDDL operators.
    In order to have  negative examples of the initiation set classifier for each operator, this operation utilizes the initiation dataset $ID$  considering all the samples with option $o$ and $f(s,o) = False$.
    The positive examples comprise instead the initial states $s$ taken from $TD$ tuples belonging to the same partition.
    The initiation set classifier of the option is computed using the Support Vector Machines (SVM) \cite{10.1023/A:1022627411411}.
    The output of this phase is the set of all the \textit{initiation set} classifiers of all the operators.
    
    \item \textbf{Effect estimation}: analogously, this step is dedicated to learning the symbols that will constitute the \textit{effects} of the PPDDL operators. The effects distribution is modelled through the Kernel density estimation \cite{10.2307/2237390,10.2307/2237880}, taking in input the final states $s'$ of each partition. 
    \item \textbf{PPDDL Domain synthesis}: finally, this step is dedicated to synthesising the PPDDL domain, characterized by the complete definition of all the operators associated with the learned options in terms of preconditions and effect symbols.
    This step entails the simple mapping of all the data structures generated during the previous steps in terms of symbolic predicates to be used as preconditions and/or effects for every operator.
\end{enumerate}

The produced PPDDL domain can be potentially used to reach any subgoal that can be expressed in terms of the available generated symbols at any point during the Discovery-Plan-Act (DPA) loop. 
One interesting aspect of the proposed DPA framework is that the semantic precision of the abstract representation and its expressiveness increase as the DPA cycles proceed, as will be described in the experimental section.

\subsubsection{Goal Selector}
\label{sub:intrinsic_motivation}
This module aims at simulating the \textit{intrinsic motivations} driving the agent towards interesting areas to satisfy its \textit{curiosity} and optimize its exploration.
In particular, one of the most fascinating aspects of this system is the capability of setting its high-level goals, which potentially could be a combination of symbols defining a state that the agent has never experienced before.
In other words, abstract reasoning could be the driving criterion for \textit{using the imagination to explore an unknown environment}.
Despite the previous goal is rather ambitious and still the object of future work, we will demonstrate in this work that the abstract reasoning can indeed be used for the more ``down-to-earth'' task of devising rational criteria to make more efficient the exploration of unexplored parts of the environment.

The selection of the target state $s_{target} \in S$ to be reached in the next exploration cycle is performed at line \ref{lst:get_target_state} of Algorithm \ref{alg:dpa_alg}, calling the procedure $Get\_Target\_State()$.
The \textit{Goal Selector} suggests such state to the system, following an internal strategy which can be, in this case, \textit{Action Babbling}, \textit{Goal Babbling} and \textit{Distance-based Goal Babbling.}
 \begin{figure}[ht]
        \centering
        \includegraphics[width=0.65\textwidth]{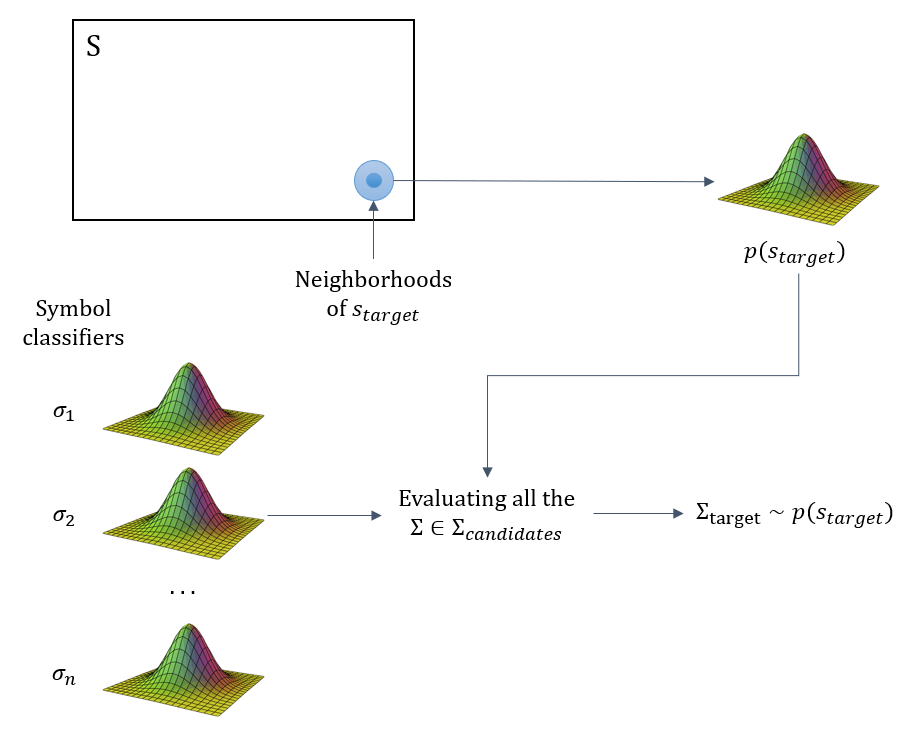}
        \caption[The conceptual depiction of the selection of the symbols necessary to represent a goal.]{The conceptual depiction of the selection of the symbols necessary to represent the goal $s_{target}$. The set of symbols having the distribution most similar to the goal will define the goal of the PPDDL problem file.}
        \label{fig:select_symbols_target}
    \end{figure}

\textit{Action Babbling} is the simplest strategy of the system for the exploration, consisting of a pure random walking of the agent.
This strategy returns $s_{target} = NULL$, so no plan $\omega^{EX}$ is generated and executed in the exploration phase on the subsequent cycle.
\textit{Goal Babbling} and \textit{Distance-based Goal Babbling} differ in the way they implement IMs (such as curiosity).
More specifically, in our system curiosity is formalised as an interest to reach (i) a random goal among those already achieved, and (ii) the border of the already acquired knowledge.

The first strategy is represented by  \textit{Goal Babbling}, consisting in randomly selecting a low-level state in the environment and trying to reach it \cite{Rolf2010goalbabbling}.
Usually, the assumption of  \textit{Goal Babbling} is that all the goals which can be set belong to the world's low-level states that are reachable; each goal is formalised as the configuration of joints or position to be reached with the robot's actuators \cite{grail_2016}.
Since in general not all the states $s \in S$ are valid (e.g. the agent can't move inside the wall), this strategy selects a random state $s_{target}$ among the visited ones (line \ref{lst:get_target_state}). 
Subsequently, $s_{target}$ is translated into a set of propositional symbols $\{\sigma_1,...,\sigma_k\}$, as described in the following subsection \ref{subsub:translating}, which represent the high-level goal to be reached using an off-the-shelf PPDDL planner.
The capability of translating low-level states into symbols gives the agent a chance to reason on causal dependencies and, consequently, plan.
It is important to notice that a pure \textit{Goal Propositional Babbling}, consisting of the selection of a random subset of high-level symbols, would not be an effective strategy because only a limited number of combinations of symbols conjunctions are valid goals.
Consequently, \textit{Goal Propositional Babbling} is not taken into consideration.

The second strategy (\textit{Distance-based Goal Babbling}) is implemented as a modified version of the \textit{Goal Babbling}, and models the curiosity towards the less explored states as being influenced by the goal's distance from its starting location $s_{init}$.
In this case, the curiosity level for a state is defined as
\begin{equation}
     \eta(s)  = \left\| s_{init} - s + \mathcal{Z} \right\|,
\end{equation}
consisting of the norm of the difference between the state vector of the agent at the beginning of each episode $s_{init}$ and the visited state $s$. 
The low-level $s_{target}$ state is selected between the farthest visited states, as follows:
\begin{equation}
    s_{target} = \max_s \eta(s),  \forall s \in S_{visited}.
\end{equation}

More precisely, the low-level state $s$, target of the exploration, is the state that maximizes the distance $\eta(s)$.
In the computation of $s_{target}$, a Gaussian noise $\mathcal{Z} \sim \mathcal{N}(0,1)$ is added, to facilitate the reaching of different states.
The \textit{Goal Selector} driven by $\eta$ continuously pushes the agent towards the border of the already acquired knowledge (similarly to the idea presented in \cite{Ecoffet2019goexplore,Bharadhwaj2021leaf}, but with a different implementation) and is thus the main responsible for the knowledge increase of the agent.
Moving towards the frontier states, the agent is more likely to encounter novel states thus maximizing the probability to acquire new symbolic knowledge, which can in its turn be used to synthesize new (e.g., more expressive) high-level goals to be eventually reached through planning, ultimately creating a virtuous loop in which the agent's capabilities of increasing its knowledge through environment exploration are iteratively enhanced.

\subsubsection{Translating low-level states into symbols}\label{subsub:translating}

The peculiarity and, simultaneously, the biggest challenge of this software architecture is thus to use an abstract symbolic representation to manage the evolution of the agent's knowledge.
Using a symbolic representation to describe a desired environment configuration is a powerful tool for an efficient exploration of the world.

After selecting $s_{target}$ (line \ref{lst:get_target_state}), the purpose of the planning module is twofold.
On the one hand, it can be employed as usual to verify the reachability of the goal $s_g$ of the environment (i.e. get the treasure and bring it "home" in the Treasure Game, in line \ref{lst:pddl_validity}) and, on the other hand, it can be exploited to generate a plan driving the agent towards states, in our case $s_{target}$, relevant to extend the knowledge of the system (line \ref{lst:planning}).
In both cases, to take advantage of planning it is necessary to transform a low-level state into a high-level one.
This operation requires finding the combination of propositional symbols
\begin{equation}
    \hat{\Sigma}_{target} = \{\sigma_1,...,\sigma_k\} 
\end{equation}
that best represent the portion of state space including $s_{target}$.
In other words, we look for a subset of symbols $\hat{\Sigma}_{target}$ whose \textit{grounding} is $s_{target}$.
These symbols make it possible to generate the definition of a planning problem (line \ref{lst:pddl_problem}) which, together with the domain definition, can be used to perform planning and solve the problem (line \ref{lst:planning}).

In order to select the right symbols conjunction, the system creates the classifier $Cl_{target} \sim p(s_{target})$ approximating a distribution over $s_{target}$.
$Cl_{target}$ is a SVM classifier trained on the states
\begin{equation}
    \{ s  | \left\| s_{target} - s \right\| \leq \epsilon \}, \forall s \in S_{visited},
\end{equation}
representing, as positive samples, all the neighbours of $s_{target}$ within a maximal distance $\epsilon$  and,  as negative samples, all the remaining states encountered in the agent's experience.
Once the classifier is generated, all the propositional symbols whose \textit{mask} is equal to a \textit{factor}\footnote{\textit{"Sets of low-level state variables that, if changed by an option execution, are always changed simultaneously"~\cite{konidaris2018skills}}.} contained by $Cl_{target}$ are collected as candidates $\hat{\Sigma}_{candidates}$.
Then, all the subsets of symbols $\hat{\Sigma}_i \subset \hat{\Sigma}_{candidates}$, \textit{whose respective masks do not overlap} are evaluated as representations of $s_{target}$.
From each $\hat{\Sigma}_i$, $m$ state samples $S_{\Sigma_i} = \{s_1,...,s_m\}$ are generated and the score of the subset $\hat{\Sigma_i}$ is calculated as
\begin{equation}
    score(\Sigma_i) = \frac{1}{m}  \sum_{s \in S_{\Sigma}} Cl_{target}(s)
\end{equation}
where $Cl_{target}(s)$ returns the probability that $s$ belongs to the positive class of the classifier $Cl_{target}$.
Then, the subset of symbols $\hat{\Sigma}$ maximizing the $score$ function is used as a goal in the problem definition $\mathcal{P}_{target}$. 

\subsubsection{Planning}
At the end of each cycle, the planning process generates a plan to reach either $s_{target}$ and $s_g$.
In both cases, in order to create a PDDL problem $\mathcal{P}$, it is necessary to find the set of symbols $\hat{\Sigma}_{init}, \hat{\Sigma}_{g}$ and $\hat{\Sigma}_{target}$, describing the most suitable high-level state representation for $s_{init}, s_g$ and $s_{target}$ respectively, as described in the previous subsection \ref{subsub:translating}.
Indeed, the couples $(\hat{\Sigma}_{init}, \hat{\Sigma}_{g})$ and $(\hat{\Sigma}_{init}, \hat{\Sigma}_{target})$ define the problems $\mathcal{P}_g$ and $\mathcal{P}_{target}$.
At line \ref{lst:pddl_problem} of Algorithm~\ref{alg:dpa_alg}, $\mathcal{P}_{target}$ is generated as previously discussed and the plan to solve it, $\omega^{EX}$, is generated by the planner (line \ref{lst:planning}).

Before moving to the next cycle, the system tries to solve also the problem $\mathcal{P}_{g}$, performing the function $Check\_PPDDL\_Validity$ (line~\ref{lst:pddl_validity}).
The resulting plan $\omega^{g}$ is only used internally by the system to keep track of the success ratio of the planner with the evolution of the synthesized knowledge of the agent.

\section{Experiment and Results}\label{sec4}
This section, dedicated to the experimental analysis, will first describe the dynamics of the environment, followed by an example of the system's cycle execution with its outputs and, finally, the overall results collected over different environment configurations. 

\begin{figure}[ht]
     \centering
     \begin{subfigure}[b]{0.325\textwidth}
         \centering
         \includegraphics[width=\textwidth]{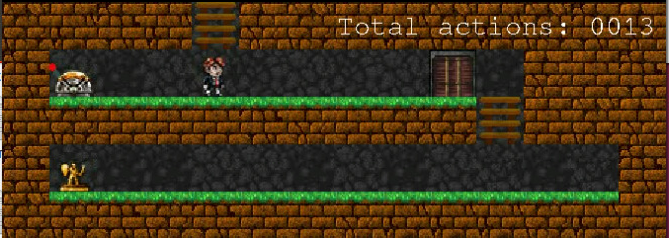}
         \caption{$domain 1$}
         \label{fig:01_domain}
     \end{subfigure}
     \hfill
     \begin{subfigure}[b]{0.325\textwidth}
         \centering
         \includegraphics[width=\textwidth]{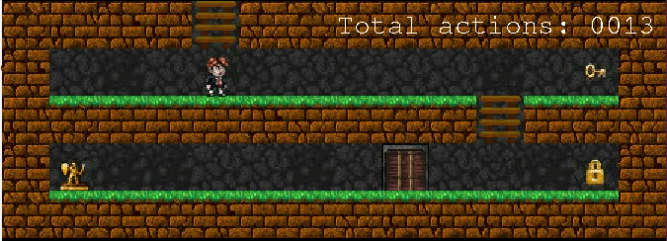}
         \caption{$domain 2$}
         \label{fig:02_domain}
     \end{subfigure}
     \hfill
     \begin{subfigure}[b]{0.325\textwidth}
         \centering
         \includegraphics[width=\textwidth]{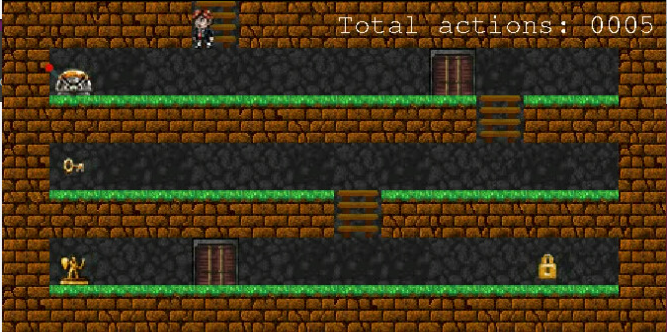}
         \caption{$domain 3$}
         \label{fig:03_domain}
     \end{subfigure}

    \begin{subfigure}[b]{0.325\textwidth}
         \centering
         \includegraphics[width=\textwidth]{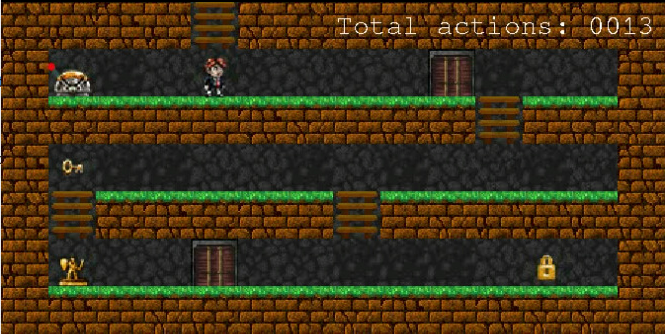}
         \caption{$domain 4$}
         \label{fig:04_domain}
     \end{subfigure}
     \begin{subfigure}[b]{0.325\textwidth}
         \centering
         \includegraphics[width=\textwidth]{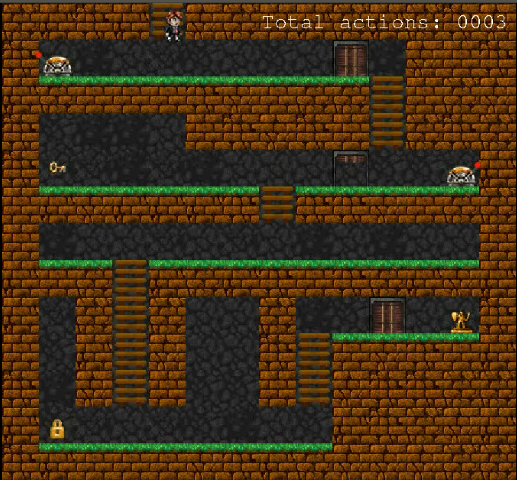}
         \caption{$domain 5$}
         \label{fig:treasure_game_domain}
     \end{subfigure}
     \hfill

        \caption[All the domain configurations used in the experimental analysis.]{All the domain configurations used in the experimental analysis. 
        The game's purpose is to get the treasure at the bottom of the maze and bring it back to the top ladder.}
        \label{fig:domains}
\end{figure}

\subsection{Environment setup}
The implemented system has been tested in the so-called Treasure Game domain \cite{konidaris2018skills}.
In such an environment, an agent can explore the maze-like space by moving through corridors and doors, climbing stairs, interacting with handles (necessary to open/close the doors), bolts, keys (necessary to unlock the bolts) and a treasure.
The agent starts its activity from the ladder on top of the maze (home location) and its overall task is to find the treasure and bring it back to the starting location.
In our experimentation, the agent starts endowed with no previous knowledge about the possible actions that can be executed in the environment; the agent is only aware of the basic motion primitives at its disposal $A = \{ go\_up, go\_down, go\_left, go\_right,interact \}$, respectively used to move the agent up, down, left or right by 2-4 pixels (the exact value is randomly selected with a uniform distribution) and to interact with the closest object.
The interaction with a lever changes the state (open/close) of the doors associated with that lever (both on the same floor or on different floors) while the interaction with the key and/or the treasure simply collects the key and/or the treasure inside the agent's bag (in the bottom-right corner of the screen).
The interaction with the bolt opens the door next to the treasure and it is feasible only when the agent has the key in its bag.
The state $s \in S$ is defined in terms of the following low-level variables:
\begin{equation}
    s = (x_{agent},y_{agent},\theta_1,\theta_2,...,x_{key},y_{key},x_{bolt},x_{treasure},y_{treasure})\label{eq:state}
\end{equation}
in which $x_{agent},y_{agent}$ is the (x,y) position of the agent, $\theta_i$ is the angle of the lever $i$, $x_{key},y_{key}$ is the (x,y) location of the key, $x_{bolt}$ is the state of the bolt (1 if open and 0 if locked) and $x_{treasure},y_{treasure}$ is the (x,y) location of the treasure.

The system is tested in five different configurations of the environment, of increasing complexity: two small-sized (Figure \ref{fig:01_domain} and \ref{fig:02_domain}), two medium-sized (Figure \ref{fig:03_domain} and \ref{fig:04_domain}) and one complete instance (Figure \ref{fig:treasure_game_domain}).
For example, in the setting depicted in Figure \ref{fig:treasure_game_domain} the agent has to: (i) pull the two levers on the top of the maze to open the doors, (ii) get the key which is used to (iii) unlock the bolt, (iv) get the treasure and (v) bring it back on top of the environment.
The obstacles that pertain to the other configurations are shown in Table \ref{tab:obstacles_table}.

\begin{table}[ht]
    \centering
    \begin{tabular}{c|c|c|c|c}
        Domain &  Levers & Keys & Bolts & Notes\\
        \hline
        $domain 1$ & 1 & 0 & 0 & None.\\
        $domain 2$ & 0 & 1 & 1 & None.\\
        $domain 3$ & 1 & 1 & 1 & None.\\
        $domain 4$ & 1 & 1 & 1 & Shortcut available.\\
        $domain 5$ & 3 & 1 & 1 & \shortstack{\\2 levers going to the treasure,\\1 going home.}\\
    \end{tabular}
    \caption{Obstacles to be solved to end the game.}
    \label{tab:obstacles_table}
\end{table}


\subsection{The cycle}
For exemplificatory purposes, a complete execution cycle of the system is briefly described in the following, showing the output of each phase of an intermediate cycle (cycle n. 10) performed on $domain 3$ (see Figure \ref{fig:03_domain}) in the \textit{Goal Babbling} strategy case. 

\paragraph{Option Generation}
At the beginning of each cycle, the agent executes Algorithm \ref{alg:discovery_alg} to collect some options exploiting the agent's primitives, before the exploration can commence.
In the Treasure Game environment selected for this work, the agent executes $d\_eps = 1$ episodes, composed by $d\_steps = 200$ primitive actions.
After the execution of the plan $\omega^{EX}$, the random exploration is reprised using the primitives contained in $A$ (see Section~\ref{subsec:options}).
The result is the following set of learned options ($11$ in total) following the formalization (\ref{eq:option_simplified}):
\begin{small}
\begin{verbatim}

O = { (go_up,{}), (go_down,{}), (go_left,{}), (go_left,go_up), 
      (go_left, go_down), (go_left,interact), (go_right,{}),
      (go_right,go_up), (go_right,go_down), (go_right,interact),
      (interact,{}) }.

\end{verbatim}
\end{small}
It is important to note that, in general, the discovered options are not all the options that may be possibly discovered in the environment, but only those experienced by the agent during the exploration.
This procedure is incremental, adding options to the set $O$ each iteration of the Algorithm \ref{alg:dpa_alg}.
As described in section \ref{subsec:option_discovery}, this procedure leverages IMs \textit{at low level}, capturing the curiosity of the agent when it discovers to have new available primitives to exploit.


\paragraph{Exploration}
In this step, the plan $\omega^{EX}$ is again executed before starting the random walk over the available options $O$.
In this particular example, 600 data entries have been collected in the $ID$ dataset and 6600 in the $TD$ dataset.
It is important to remember that the number of collected data over the cycles is not constant because when an option does not produce effects on the environment, no data is collected.

\begin{figure}
\begin{scriptsize}
\begin{verbatim}
(define (domain TreasureGame)
    (:requirements :strips :probabilistic-effects :rewards)

    (:predicates
        (notfailed)
        (symbol_0)
        (symbol_1)
        (symbol_2)
        ...
        (symbol_25)
    )

    (:action option-0-partition-0-0
        :parameters ()
        :precondition (and (notfailed) (symbol_5) (symbol_22) 
                      (symbol_14))
        :effect (and (symbol_6) (not (symbol_5)) (decrease
                (reward) 36.00))
    )

    ...

     (:action option-10-partition-3-715
        :parameters ()
        :precondition (and (notfailed) (symbol_20) (symbol_22) 
                      (symbol_14) (symbol_15) (symbol_0))
        :effect (and (symbol_13) (not (symbol_20)) (decrease 
                (reward) 90.00))
    )
)
\end{verbatim} 
\end{scriptsize}
\caption[An extract of the PPDDL generated by the abstraction procedure.]{An extract of the PPDDL generated by the abstraction procedure. Notice that \texttt{option-0-partition-0-0} can be explained by looking at the symbols of Figure \ref{fig:symbols}. In fact, the effect is to change the x position replacing $symbol\_5$ with $symbol\_6$, maintaining invariant the presence of $symbol\_14$ and $symbol\_22$. Consequently, it is the option moving the agent on the right on the last floor.}
\label{fig:pddl_representation}
\end{figure}

\begin{figure}
    \centering
    \begin{subfigure}[b]{0.4\textwidth}
        \centering
        \includegraphics[width=\textwidth]{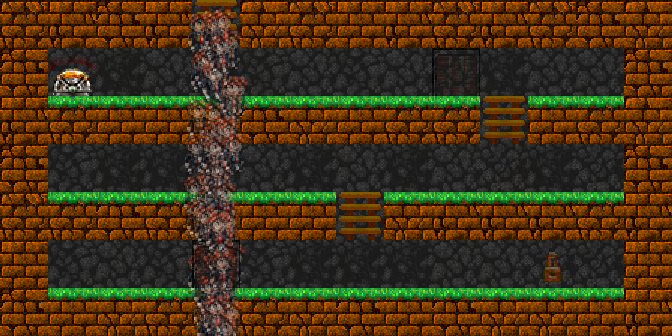}
        \caption{symbol\_5}
        \label{fig:symbol_5}
    \end{subfigure}
    \begin{subfigure}[b]{0.4\textwidth}
        \centering
        \includegraphics[width=\textwidth]{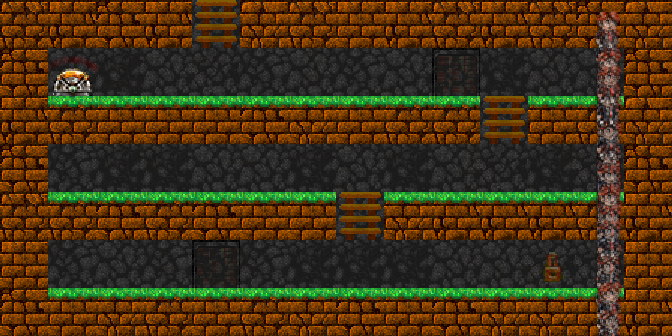}
        \caption{symbol\_6}
        \label{fig:symbol_6}
    \end{subfigure}
    \begin{subfigure}[b]{0.4\textwidth}
        \centering
        \includegraphics[width=\textwidth]{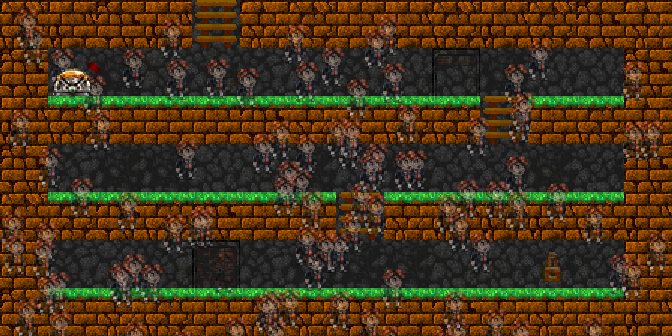}
        \caption{symbol\_14}
        \label{fig:symbol_14}
    \end{subfigure}
    \begin{subfigure}[b]{0.4\textwidth}
        \centering
        \includegraphics[width=\textwidth]{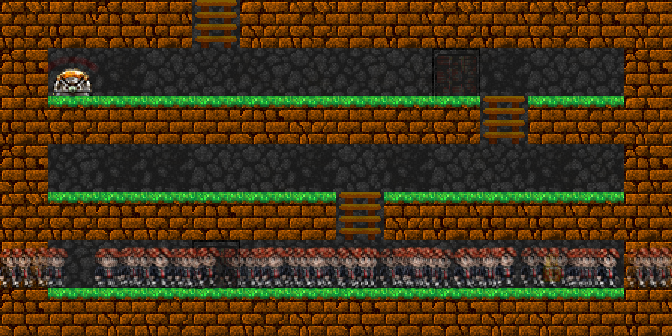}
        \caption{symbol\_22}
        \label{fig:symbol_22}
    \end{subfigure}
       \caption[Example of the symbols' semantics of a generated PPDDL's operator.]{Semantics of the symbols employed by the first operator of the generated PPDDL: (a) and (b) mean being in a certain \textit{x} position, (c) mean the lever is pulled and (d) being on the bottom floor.}
       \label{fig:symbols}
\end{figure}

\paragraph{Abstraction}
Figure \ref{fig:pddl_representation} shows a part of the output of the PPDDL domain obtained from the abstraction procedure.
As visible, the figure does present a valid PPDDL domain description, upon which the planner may reason. Moreover, any subset of the produced domain predicates (symbols) can be used to define high-level goals the planner may try to plan for.
In this specific case, the system generated 26 symbols and 716 operators.

\paragraph{Planning}
The produced PPDDL domain is used by the system to (1) solve the entire game, or task, and (2) improve the exploration.
Figure \ref{fig:planning_problems} illustrates two PDDL problems: the problem on top refers to the general goal of solving the game ($\mathcal{P}_g$) while the problem at the bottom refers to the goal dynamically synthesized by the \textit{Goal Selector} module, which in this example relates to reaching the left corner of the middle floor ($\mathcal{P}_{target}$).
The solution for both problems is presented in Figure \ref{fig:planning_solutions}.
The first solution, $\omega^g$, solves the game problem in 21 moves and the second one, $\omega^{EX}$, reaches the \textit{Goal Selector} goal in 6 moves.

\begin{figure}
\begin{scriptsize}
\begin{verbatim}
(define (problem task_goal)
    (:domain TreasureGame)

    (:init (notfailed) (symbol_0) (symbol_1) 
           (symbol_2) (symbol_3) 
           (symbol_4) (symbol_5) )

    (:goal (and (notfailed) (symbol_4) 
           (symbol_17) (symbol_18)) )
)


(define (problem im_goal)
    (:domain TreasureGame)

    (:init (notfailed) (symbol_0) 
           (symbol_1) (symbol_2)
           (symbol_3) (symbol_4) 
           (symbol_5) )

    (:goal (and (symbol_13) (symbol_24) 
           (symbol_14) (symbol_2)
           (notfailed)) )
)

\end{verbatim} 
\end{scriptsize}
\caption[Example of PPDDL problems synthesized by the system.]{On the top, the problem of solving the task is synthesized by the system $\mathcal{P}_g$ and, on the bottom, the problem generated by the \textit{Goal Selector} module $\mathcal{P}_{target}$.}
\label{fig:planning_problems}
\end{figure}

\begin{figure}
\begin{scriptsize}
\begin{verbatim}
PLAN TASK GOAL:

[ 1:(go_down,{}),           ; climb down the stairs
  2:(go_left,go_down),      ; go left until it can go down
  3:(interact,{}),          ; pull the lever
  4:(go_right,go_down),     ; go right up to the stairs to go down
  5:(go_down,{}),           ; climb down the stairs
  6:(go_right,{}),          ; go right up to the end of the corridor 
  7:(go_left,{}),           ; go left up to the end of the corridor
  8:(interact,{}),          ; take the key   
  9:(go_right,go_down),     ; go right up to the stairs
  10:(go_down,{}),          ; climb down the stairs
  11:(go_right,interact),   ; go right up to the bolt
  12:(interact,{}),         ; unlock the bolt
  13:(go_left,go_down),     ; go left until it is possibile
  14:(interact,{}),         ; get the treasure
  15:(go_right,go_up),      ; go right up to the stairs
  16:(go_up,{}),            ; go upstairs
  17:(go_right,{}),         ; go right up to the end of the corridor
  18:(go_left,go_up),       ; go left up to the stairs
  19:(go_up,{}),            ; go upstairs
  20:(go_left,go_up),       ; go left up to the stairs
  21:(go_up,{}) ]           ; go up to home

PLAN IM GOAL:

[ 1:(go_down,{}),           ; climb down the stairs
  2:(go_left,go_down),      ; go left up to the lever
  3:(interact,{}),          ; pull the lever
  4:(go_right,go_down),     ; go right until it can go down
  5:(go_down,{}),           ; climb down the stairs
  6:(go_left,interact) ]    ; go left up to the key
\end{verbatim} 
\end{scriptsize}
\caption{The plans generated $\omega^{g}$ and $\omega^{EX}$.}
\label{fig:planning_solutions}
\end{figure}

\subsection{Results}

In this section, the overall results of the system over different domain instances are described.
First, the setting of the environment is discussed, then the adopted baseline, as well as the other strategies enabling the planning exploration.
Subsequently, some charts are presented, that highlight the system's performance in terms of success ratio on the planning task.
Finally, some issues worth being underscored are commented, and the limitations of the employed technologies are discussed.

For our purposes, the Treasure Game\footnote{Github repository: https://github.com/sd-james/gym-treasure-game} has been used with five different mazes (see Figure \ref{fig:domains}), to focus on the performances of the system on smaller domains, highlighting pros and cons of the symbolic approach proposed.
The system has been executed, following the workflow described in Algorithm \ref{alg:dpa_alg}, in the cited five mazes configurations using parameters suitable to solve the tasks, which are described later.
To summarize, the system iteratively (i) looks for new options $O_{new}$ performing $d\_steps$ primitives for $d\_eps$ episodes, (ii) explores for $dpa\_eps$ episodes the maze executing $dpa\_steps$, (iii) creates a symbolic abstraction of the domain $\mathcal{D}$, (iv) creates a plan $\omega^{EX}$ to optimize the exploration of the successive cycle $c + 1$.
It is important to note that the autonomously discovered options successfully produced a valid high-level model, usable to build correct plans.
To assess the accuracy of the current abstraction $\mathcal{D}$, we evaluate its ability to reach the goal $s_g$ at the end of each cycle.
This is done by requesting the planner to solve the problem $\mathcal{P}_g$ using the currently produced symbolic domain description $\mathcal{D}$.

The first strategy used in the experiments is the random walk, which is called \textit{Action Babbling}~\cite{chitnis2021glib} (see Section~\ref{sub:intrinsic_motivation}).
This strategy simply executes the Algorithm \ref{alg:dpa_alg} without using planning because the \textit{Goal Selector} returns $s_{target} = NULL$ and no plan $\omega^{EX}$ is generated.
This simple strategy is used as a baseline for the experimental analysis, and it is necessary to fully appreciate the advantages of using the symbolic approach.
The exploration is equivalent to the one used by Konidaris \cite{konidaris2018skills}, and we use it to observe the development of the symbolic model over time.

To support the use of symbolic planning, two other strategies have been considered: \textit{Goal Babbling} and \textit{Distance-based Goal Babbling} (see Section~\ref{sub:intrinsic_motivation}).
These three strategies could be seen as having an increasing complexity and effectiveness in the exploration:
\begin{itemize}
    \item \textit{Action Babbling} selects completely random actions;
    \item \textit{Goal Babbling} selects random goals, reaching them with a planned sequence of actions;
    \item \textit{Distance-based Goal Babbling} selects the farthest goals and reaches them by using planning (see subsection \ref{sub:intrinsic_motivation}).
\end{itemize}
It is reasonable to conjecture that in smaller domains where the reasoning is less necessary and purely random exploration may suffice, we do not expect to observe much difference among the previous strategies; while in the bigger domains the time to reach the goal may significantly change, depending on the strategy employed.

The five configurations considered are depicted in Figure \ref{fig:domains}, and the parameter values used in the exploration function $Collect\_Data$ for each configuration are shown in Table \ref{tab:domains_exp_parameters}.
Smaller domains $domain 1$ and $domain 2$ required the execution of $50$ options per episode, $domain 3$ required $150$, $domain 4$ required $200$, and finally $800$ options were executed in $domain 5$.

\begin{table}[ht]
    \centering
    \begin{tabular}{c|c|c|c}
        Domain &  $dpa\_eps $ & $dpa\_steps$ & minimal solution steps\\
        \hline
        $domain 1$ & 4 & 50 & 11\\
        $domain 2$ & 4 & 50 & 13\\
        $domain 3$ & 4 & 150 & 19\\
        $domain 4$ & 4 & 200 & 15\\
        $domain 5$ & 4 & 800 & 31\\
    \end{tabular}
    \caption{Main parameters used in the different environment configurations.}
    \label{tab:domains_exp_parameters}
\end{table}

\begin{figure}
    \centering
    \begin{subfigure}[b]{0.49\textwidth}
        \centering
        \includegraphics[width=\textwidth]{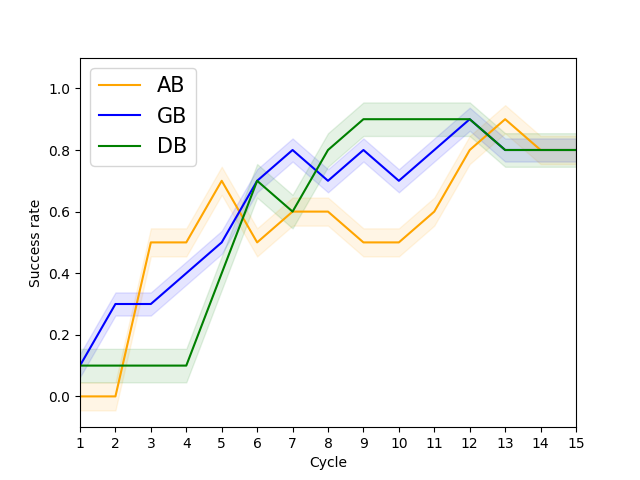}
        \caption{domain 1}
        \label{fig:planning_results_a}
    \end{subfigure}
    \hfill
    \begin{subfigure}[b]{0.49\textwidth}
        \centering
        \includegraphics[width=\textwidth]{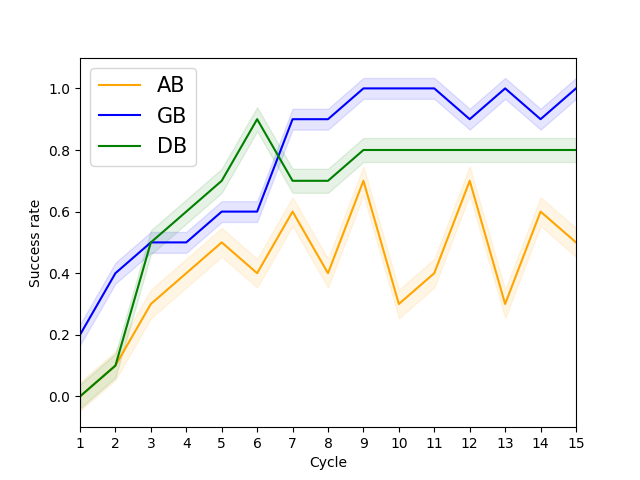}
        \caption{domain 2}
        \label{fig:planning_results_b}
    \end{subfigure}
    \begin{subfigure}[b]{0.49\textwidth}
        \centering
        \includegraphics[width=\textwidth]{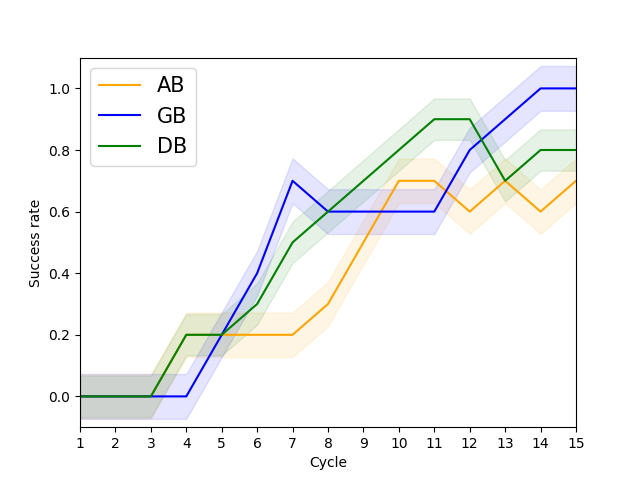}
        \caption{domain 3}
        \label{fig:planning_results_c}
    \end{subfigure}
    \hfill
    \begin{subfigure}[b]{0.49\textwidth}
        \centering
        \includegraphics[width=\textwidth]{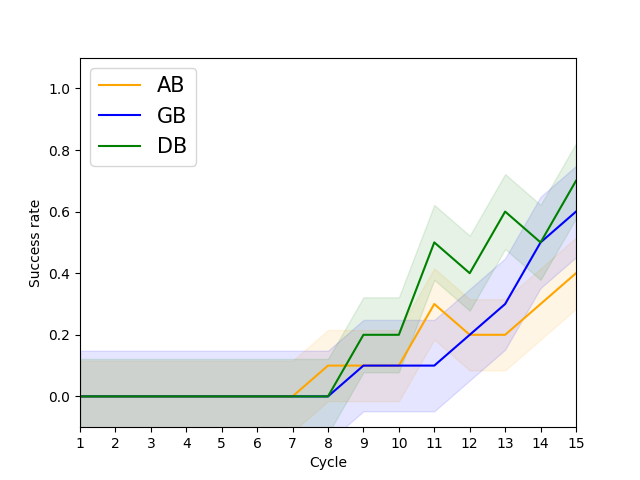}
        \caption{domain 4}
        \label{fig:planning_results_d}
    \end{subfigure}
    \begin{subfigure}[b]{0.49\textwidth}
        \centering
        \includegraphics[width=\textwidth]{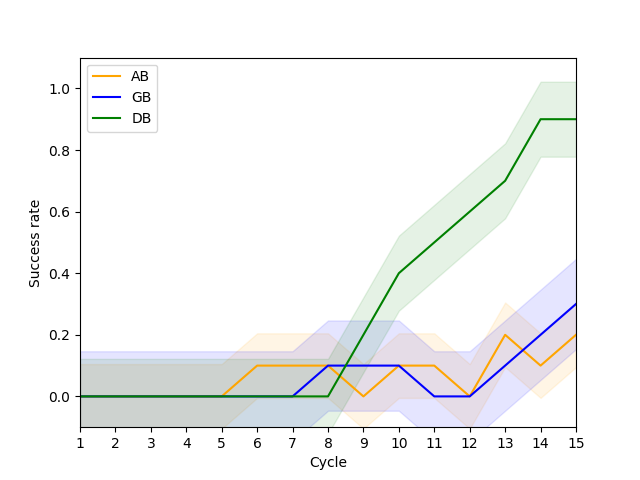}
        \caption{domain 5}
        \label{fig:planning_results_e}
    \end{subfigure}
       \caption[Success rate of $\omega^g$ using different exploration strategies over time.]{The figure depicts the success rate of $\omega^g$ using different exploration strategies over time. In figure \ref{fig:planning_results_a}, \ref{fig:planning_results_b}, \ref{fig:planning_results_c}, \ref{fig:planning_results_d} and \ref{fig:planning_results_e}, we have respectively the results of the domain of figures \ref{fig:01_domain}, \ref{fig:02_domain}, \ref{fig:03_domain}, \ref{fig:04_domain}, \ref{fig:treasure_game_domain} over 15 cycles of exploration.}
       \label{fig:planning_results}
\end{figure}

The main results of the system are depicted in Figure \ref{fig:planning_results}.
Precisely, in the graphs, it is shown the probability of success in solving the game over time using different strategies.
Such success entails the generation of a sufficiently mature domain $\mathcal{D}$ and a correct problem formalization of $\mathcal{P}_{g}$, resulting in a correct plan $\omega^g$.

The system has been run with $cycles = 15$, assessing at the end of each cycle whether $\omega^g$ was able to solve the game using the current synthesized PPDDL domain representation $\mathcal{D}$.
This mechanism has been performed for ten trials.
Consequently, in the charts of Figure \ref{fig:planning_results} are depicted cycles on the \textit{x-axis}, and the percentage of trials that have succeeded in solving the game in the specific cycle on the \textit{y-axis}.

In the smaller domains $domain 1$, $domain 2$, $domain 3$ and $domain 4$, the planning contribution is limited and sometimes not visible at all.
Especially in the simplest domain $domain 1$ the three strategies present similar performances, solving in the last cycles 80-90\% of the trials (Figure \ref{fig:planning_results_a}).
In domains $domain 2$ and $domain 3$, presenting slightly higher complexities, the advantages of executing plans start being visible (Figure \ref{fig:planning_results_b} and \ref{fig:planning_results_c}).
It is evident that \textit{Goal Babbling} behaves almost perfectly in the last cycles, demonstrating the usefulness of collecting transition data with reasonable sequences of actions.
The combination of randomness in selecting the goal to be reached and the reasoned transitions speed up the exploration and the consequent maturity of the PPDDL representation of the environment.
Instead, the \textit{Distance-based Goal Babbling} seems better than a completely random search but less effective than the precedent.
This result entails that applying an exploration focused on the frontier's goals is not convenient in smaller domains.
In fact, in the conclusive cycles of $domain 2$ and $domain 3$, \textit{Goal Babbling} maintains 90-100\% of success ratio, \textit{Distance-based Goal Babbling} around 80\% and \textit{Action Babbling} between 60-70\%.

Although the problem faced by $domain 3$ seems easier than $domain 4$, presenting respectively minimal solutions of 19 and 15 steps (see Table \ref{tab:domains_exp_parameters}), the system needs more steps per episode to solve $domain 4$ using the same amount of episodes used by $domain 3$.
The reason of this behaviour is due to the higher complexity of synthesizing a PPDDL domain $\mathcal{D}$ for this scenario.
In fact, the scenario $domain 4$ presents a sort of "shortcut" to reach the treasure, which does not require collecting the key and opening the door.
From the point of view of the abstraction procedure, the shortcut represents a branch in the possible choices of the agent, thus presenting the agent with additional concepts to be abstracted in order to synthesize a complete representation.
Consequently, the system generates additional symbols and operators, requiring more experience to strengthen the transition model captured by the PPDDL and provide satisfying plans.

The significance of the frontier exploration emerges in the $domain 5$ configuration,  where planning evidently boosts the results of the agent.
Indeed, being bigger than other domains, $domain 5$ is more difficult to be solved and planning results to be efficient in driving the exploration of the agent immediately towards the borders of its knowledge.
The main result highlighted by the charts is that in bigger domains (Figure \ref{fig:planning_results_e}), the impact of using planning is evident.
Planning is able to easily drive the agent towards interesting visited states, where it can continue exploring.
Statistically, after collecting the transition data for 15 cycles, the system struggles to solve the problem, never exceeding 10\% of success.
Similar behaviour for \textit{Goal Babbling}, reaching most 20\% of success.
Instead, \textit{Distance-based Goal Babbling} results being extremely effective, reaching 90\% of success.
Conceptually, the strategy continuously pushes the agent towards the frontiers of the visited states, increasing the probability of encountering new unexplored areas and reducing redundant data.

\subsection{Discussion and Future Work}

As shown in Figure \ref{fig:planning_results}, the results are not deterministic and change over time.
On the one hand, the system is continuously evolving in terms of knowledge and, on the other hand, ML techniques introduce further stochasticity to the final representation generated.
During the abstraction procedure, described in section \ref{subsec:abstraction}, some statistical tools are employed to create data structures to represent preconditions, effects and symbols.

\begin{figure}[ht]
        \centering
        \includegraphics[scale=0.2]{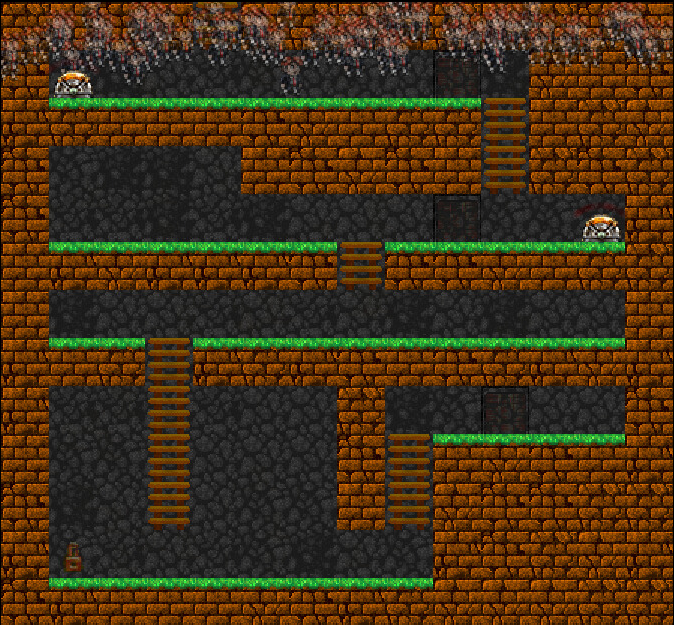}
        \caption[The graphical representation of a potentially noisy symbol.]{The graphical representation of the symbol "on the higher y-axis position". It can be seen that it is significantly noisy because some samples of the agent are depicted on the top y-axis and other samples almost on the lower floor.}
        \label{fig:top_ladder_symbol}
\end{figure}

For instance, an erroneous\footnote{Statistical methods are not mistaken because they just create a representation based on the data provided. However, for our purposes, some models' instances could be obstacle on the reaching of our goal.} clustering phase could generate unexpected effects on symbols and operators.
An example highlighting this fact is the \textit{noisiness of some symbols}, interfering with the generation of a correct formalization of $\mathcal{D}, \mathcal{P}$ and, consequently, the resulting plan $\omega$.
In Figure \ref{fig:top_ladder_symbol} it can be seen the graphical representation of the symbol "on the highest y-axis position", meaning "on the top ladder" because it is the only possible state with such y-axis value (it is not allowed to move inside the walls).
In some cases, it could happen that the initial state of the game is interpreted as being at the top floor under the ladder and, consequently, it is not necessary to climb down the ladder to execute the agent's task.
Then, although almost the whole plan $\omega$ is correct to complete the game, without the option of climbing down the ladder as the first action, the plan is incomplete, and the trial is considered unsuccessful.
On the one hand, this is a drawback of using ML tools, which can introduce noise in the symbolic representation.
On the other hand, it is the strong point of the probabilistic approach, always proposing solutions, even though presenting a certain degree of error.

The system proposed is limited by the \textit{propositional logic} representation, which does not permit the inclusion of arguments in operators and symbols.
Such limitation could be overtaken by developing a new abstraction procedure following other logics (e.g. First Order Logic (FOL)).
Recently, an implementation of an object-centric abstraction procedure has been proposed \cite{james2021autonomous}, demonstrating that it is possible to create a lifted representation suitable for transfer the knowledge to new tasks.

Another aspect to be analyzed deeper is the \textit{management of the knowledge data}.
Unfortunately, the abstraction module has no real incremental nature, but the abstraction procedure is executed over all the data, each cycle from scratch.
This way of operating is not computationally efficient and does not reflect the learning process of the human being.
However, given that in our domains all the necessary knowledge is discovered in a certain amount of time and does not change, a first improvement to be applied to the system could be maintaining a maximum of transition and initiation data discarding over-sampled transitions.
Therefore, a sort of filter in the knowledge acquisition could stop the growth in terms of the abstraction procedure's time in stationary and limited domains but also reduce it for all the other cases.

Then, the abstraction time seems being linear with respect to the number transition tuples (in the worst-case exponential according to Konidaris \cite{konidaris2018skills}).
Consequently, it would be fundamental to find solutions to mitigate this aspect.
Possible actions could be to filter the acquired data or structure the collected knowledge in another form more efficiently for the abstraction process and, eventually, produce different complementary representations which are used according to the task to be tackled.

Finally, a natural evolution of this work lies in the neuro-symbolic approach.
The integration of explicit knowledge inside sub-symbolic systems makes the learning process more effective and efficient, taking advantage of both the approaches.
One of most promising extension of this work should embrace this new techniques, as some other preliminar works have already done \cite{Asai_Fukunaga_2018,GARNELO201917, garnelo2016towards}.

\section{Conclusions}\label{sec5}
In this paper, a novel approach for open-ended learning in autonomous agents based on intrinsically motivated planning is presented. 
This approach integrates two powerful paradigms, intrinsic motivation and classical planning, to enable agents to continuously learn and improve their knowledge and skills without relying on external supervision or rewards.

This work suggests an alternative or complementary approach to the advanced and popular sub-symbolic methods, demonstrating interesting features. 
First, it allows agents to explore and learn in a self-directed and open-ended manner without being limited to a predefined set of goals or tasks. 
Second, it enables agents to represent and reason about their knowledge and skills in a structured and formal way, which can facilitate planning and generalization to new situations. 
Third, it can incorporate intrinsic motivations that drive the agent to explore and learn beyond extrinsic goals, which can enhance the agent's adaptability, robustness, and creativity.
However, there are still several challenges and opportunities for future research in this area to enable the systems to perform complex activities in a relevant operational environment.

Overall, we believe that our approach represents a promising step towards more autonomous and intelligent agents that can continuously learn and improve in an open-ended and self-directed manner.

\backmatter




% ...

\bmhead{Conflict of Interest}
The authors declare that they have no conflicts of interest.

\bmhead{Data availability}
The implementation of the system is available at \href{https://github.com/gabrielesartor/discover_plan_act}{https://github.com/gabrielesartor/discover\_plan\_act}.


\bibliography{amiede_Publications,bibliography}

\begin{thebibliography}{10}
\providecommand{\doi}[1]{\url{https://doi.org/#1}}
\bibcommenthead

\bibitem[\protect\citeauthoryear{Campbell et~al.}{2002}]{deepblue2002}
Campbell M, Hoane AJ, hsiung Hsu F.
\newblock Deep Blue.
\newblock Artificial Intelligence. 2002;134(1):57--83.
\newblock \doi{https://doi.org/10.1016/S0004-3702(01)00129-1}.

\bibitem[\protect\citeauthoryear{Silver et~al.}{2018}]{silver2018alphazero}
Silver D, Hubert T, Schrittwieser J, Antonoglou I, Lai M, Guez A, et~al.
\newblock A general reinforcement learning algorithm that masters chess, shogi, and Go through self-play.
\newblock Science. 2018;362(6419):1140--1144.

\bibitem[\protect\citeauthoryear{Silver et~al.}{2016}]{silver2016mastering}
Silver D, Huang A, Maddison CJ, Guez A, Sifre L, Van Den~Driessche G, et~al.
\newblock Mastering the game of Go with deep neural networks and tree search.
\newblock nature. 2016;529(7587):484--489.

\bibitem[\protect\citeauthoryear{Silver et~al.}{2017}]{silver2017mastering}
Silver D, Schrittwieser J, Simonyan K, Antonoglou I, Huang A, Guez A, et~al.
\newblock Mastering the game of go without human knowledge.
\newblock nature. 2017;550(7676):354--359.

\bibitem[\protect\citeauthoryear{Oh et~al.}{2015}]{oh2015action}
Oh J, Guo X, Lee H, Lewis RL, Singh S.
\newblock Action-conditional video prediction using deep networks in atari games.
\newblock Advances in neural information processing systems. 2015;28.

\bibitem[\protect\citeauthoryear{Berner et~al.}{2019}]{berner2019dota}
Berner C, Brockman G, Chan B, Cheung V, D{\k{e}}biak P, Dennison C, et~al.
\newblock Dota 2 with large scale deep reinforcement learning.
\newblock arXiv preprint arXiv:191206680. 2019;.

\bibitem[\protect\citeauthoryear{Koch et~al.}{2019}]{koch2019reinforcement}
Koch W, Mancuso R, West R, Bestavros A.
\newblock Reinforcement learning for UAV attitude control.
\newblock ACM Transactions on Cyber-Physical Systems. 2019;3(2):1--21.

\bibitem[\protect\citeauthoryear{Nguyen and La}{2019}]{nguyen2019review}
Nguyen H, La H.
\newblock Review of deep reinforcement learning for robot manipulation.
\newblock In: 2019 Third IEEE International Conference on Robotic Computing (IRC). IEEE; 2019. p. 590--595.

\bibitem[\protect\citeauthoryear{Ibarz et~al.}{2021}]{ibarz2021train}
Ibarz J, Tan J, Finn C, Kalakrishnan M, Pastor P, Levine S.
\newblock How to train your robot with deep reinforcement learning: lessons we have learned.
\newblock The International Journal of Robotics Research. 2021;40(4-5):698--721.

\bibitem[\protect\citeauthoryear{Bellemare et~al.}{2020}]{bellemare2020autonomous}
Bellemare MG, Candido S, Castro PS, Gong J, Machado MC, Moitra S, et~al.
\newblock Autonomous navigation of stratospheric balloons using reinforcement learning.
\newblock Nature. 2020;588(7836):77--82.

\bibitem[\protect\citeauthoryear{Sutton and Barto}{2018}]{sutton2018reinforcement}
Sutton RS, Barto AG.
\newblock Reinforcement learning: An introduction.
\newblock MIT press; 2018.

\bibitem[\protect\citeauthoryear{Bengio et~al.}{2017}]{bengio2017deep}
Bengio Y, Goodfellow I, Courville A.
\newblock Deep learning. vol.~1.
\newblock MIT press Cambridge, MA, USA; 2017.

\bibitem[\protect\citeauthoryear{Butler}{1953}]{butler1953discrimination}
Butler RA.
\newblock Discrimination learning by rhesus monkeys to visual-exploration motivation.
\newblock Journal of Comparative and Physiological Psychology. 1953;46(2):95.

\bibitem[\protect\citeauthoryear{Harlow}{1950}]{harlow1950learning}
Harlow HF.
\newblock Learning and satiation of response in intrinsically motivated complex puzzle performance by monkeys.
\newblock Journal of comparative and physiological psychology. 1950;43(4):289.

\bibitem[\protect\citeauthoryear{Montgomery}{1954}]{montgomery1954role}
Montgomery KC.
\newblock The role of the exploratory drive in learning.
\newblock Journal of Comparative and Physiological Psychology. 1954;47(1):60.

\bibitem[\protect\citeauthoryear{Berlyne}{1950}]{berlyne1950novelty}
Berlyne DE.
\newblock Novelty and curiosity as determinants of exploratory behaviour.
\newblock British journal of psychology. 1950;41(1):68.

\bibitem[\protect\citeauthoryear{Berlyne}{1966}]{berlyne1966curiosity}
Berlyne DE.
\newblock Curiosity and Exploration: Animals spend much of their time seeking stimuli whose significance raises problems for psychology.
\newblock Science. 1966;153(3731):25--33.

\bibitem[\protect\citeauthoryear{Ryan and Deci}{2000}]{RYAN2000motivantion}
Ryan RM, Deci EL.
\newblock Intrinsic and Extrinsic Motivations: Classic Definitions and New Directions.
\newblock Contemporary Educational Psychology. 2000;25(1):54 -- 67.
\newblock \doi{https://doi.org/10.1006/ceps.1999.1020}.

\bibitem[\protect\citeauthoryear{D{\"u}zel et~al.}{2010}]{duzel2010novelty}
D{\"u}zel E, Bunzeck N, Guitart-Masip M, D{\"u}zel S.
\newblock NOvelty-related motivation of anticipation and exploration by dopamine (NOMAD): implications for healthy aging.
\newblock Neuroscience \& Biobehavioral Reviews. 2010;34(5):660--669.

\bibitem[\protect\citeauthoryear{Ranganath and Rainer}{2003}]{ranganath2003neural}
Ranganath C, Rainer G.
\newblock Neural mechanisms for detecting and remembering novel events.
\newblock Nature Reviews Neuroscience. 2003;4(3):193--202.

\bibitem[\protect\citeauthoryear{Redgrave and Gurney}{2006}]{redgrave2006short}
Redgrave P, Gurney K.
\newblock The short-latency dopamine signal: a role in discovering novel actions?
\newblock Nature reviews neuroscience. 2006;7(12):967--975.

\bibitem[\protect\citeauthoryear{Santucci et~al.}{2020}]{Santucci2020}
Santucci VG, Oudeyer PY, Barto A, Baldassarre G.
\newblock Intrinsically motivated open-ended learning in autonomous robots.
\newblock Frontiers in neurorobotics. 2020;13:115.

\bibitem[\protect\citeauthoryear{Oudeyer et~al.}{2007}]{Oudeyer2007intrinsic}
Oudeyer PY, Kaplan F, Hafner V.
\newblock Intrinsic motivation systems for autonomous mental development.
\newblock IEEE transactions on evolutionary computation. 2007;11(2):265--286.

\bibitem[\protect\citeauthoryear{Baldassarre and Mirolli}{2013}]{Baldassarre2013Book}
Baldassarre G, Mirolli M.
\newblock Intrinsically Motivated Learning in Natural and Artificial Systems.
\newblock Springer Science \& Business Media; 2013.

\bibitem[\protect\citeauthoryear{Frank et~al.}{2014}]{Frank2014}
Frank M, Leitner J, Stollenga M, F{\"o}rster A, Schmidhuber J.
\newblock Curiosity driven reinforcement learning for motion planning on humanoids.
\newblock Frontiers in neurorobotics. 2014;7:25.

\bibitem[\protect\citeauthoryear{Bellemare et~al.}{2016}]{Bellemare2016}
Bellemare M, Srinivasan S, Ostrovski G, Schaul T, Saxton D, Munos R.
\newblock Unifying count-based exploration and intrinsic motivation.
\newblock Advances in neural information processing systems. 2016;29.

\bibitem[\protect\citeauthoryear{Santucci et~al.}{2016}]{grail_2016}
Santucci VG, Baldassarre G, Mirolli M.
\newblock GRAIL: A Goal-Discovering Robotic Architecture for Intrinsically-Motivated Learning.
\newblock IEEE Transactions on Cognitive and Developmental Systems. 2016;8(3):214--231.
\newblock \doi{10.1109/TCDS.2016.2538961}.

\bibitem[\protect\citeauthoryear{Colas et~al.}{2019}]{Colas2019}
Colas C, Fournier P, Chetouani M, Sigaud O, Oudeyer PY.
\newblock CURIOUS: intrinsically motivated modular multi-goal reinforcement learning.
\newblock In: International conference on machine learning. PMLR; 2019. p. 1331--1340.

\bibitem[\protect\citeauthoryear{Blaes et~al.}{2019}]{blaes2019control}
Blaes S, Vlastelica~Pogan{\v{c}}i{\'c} M, Zhu J, Martius G.
\newblock Control what you can: Intrinsically motivated task-planning agent.
\newblock Advances in Neural Information Processing Systems. 2019;32.

\bibitem[\protect\citeauthoryear{Sigaud et~al.}{2023}]{Sigaud2023OEL}
Sigaud O, Baldassarre G, Colas C, Doncieux S, Duro R, Perrin-Gilbert N, et~al.
\newblock A Definition of Open-Ended Learning Problems for Goal-Conditioned Agents.
\newblock arXiv preprint arXiv:231100344. 2023;.

\bibitem[\protect\citeauthoryear{Baldassarre et~al.}{2024}]{Baldassarre2024Purpose}
Baldassarre G, Duro RJ, Cartoni E, Khamassi M, Romero A, Santucci VG.
\newblock Purpose for Open-Ended Learning Robots: A Computational Taxonomy, Definition, and Operationalisation.
\newblock arXiv preprint arXiv:240302514. 2024;.

\bibitem[\protect\citeauthoryear{Singh et~al.}{2004}]{SinghBartoChentanez2004}
Singh S, Barto AG, Chentanez N.
\newblock Intrinsically Motivated Reinforcement Learning.
\newblock In: Proceedings of the 17th International Conference on Neural Information Processing Systems. NIPS'04. Cambridge, MA, USA: MIT Press; 2004. p. 1281–1288.

\bibitem[\protect\citeauthoryear{Forestier et~al.}{2017}]{Forestier2017}
Forestier S, Portelas R, Mollard Y, Oudeyer PY.
\newblock Intrinsically motivated goal exploration processes with automatic curriculum learning.
\newblock arXiv preprint arXiv:170802190. 2017;.

\bibitem[\protect\citeauthoryear{Romero et~al.}{2022}]{Romero2022Curriculum}
Romero A, Baldassarre G, Duro RJ, Santucci VG.
\newblock Autonomous learning of multiple curricula with non-stationary interdependencies.
\newblock In: 2022 IEEE International Conference on Development and Learning (ICDL). IEEE; 2022. p. 272--279.

\bibitem[\protect\citeauthoryear{Bengio et~al.}{2009}]{Bengio2009}
Bengio Y, Louradour J, Collobert R, Weston J.
\newblock Curriculum learning.
\newblock In: Proceedings of the 26th annual international conference on machine learning; 2009. p. 41--48.

\bibitem[\protect\citeauthoryear{Machado et~al.}{2017}]{Machado2017}
Machado MC, Bellemare MG, Bowling M.
\newblock A laplacian framework for option discovery in reinforcement learning.
\newblock arXiv preprint arXiv:170300956. 2017;.

\bibitem[\protect\citeauthoryear{Oddi et~al.}{2020}]{oddi2020integrating}
Oddi A, Rasconi R, Santucci VG, Sartor G, Cartoni E, Mannella F, et~al.
\newblock Integrating open-ended learning in the sense-plan-act robot control paradigm.
\newblock In: ECAI 2020. IOS Press; 2020. p. 2417--2424.

\bibitem[\protect\citeauthoryear{Sartor et~al.}{2021}]{SartorZMORS21}
Sartor G, Zollo D, Mayer MC, Oddi A, Rasconi R, Santucci VG.
\newblock Autonomous Generation of Symbolic Knowledge via Option Discovery.
\newblock In: Proceedings of the 9th Italian workshop on Planning and Scheduling (IPS'21) and the 28th International Workshop on "Experimental Evaluation of Algorithms for Solving Problems with Combinatorial Explosion" (RCRA'21). vol. 3065 of {CEUR} Workshop Proceedings. CEUR-WS.org; 2021. .

\bibitem[\protect\citeauthoryear{Sartor et~al.}{2022}]{sartor2022option}
Sartor G, Zollo D, Cialdea~Mayer M, Oddi A, Rasconi R, Santucci VG.
\newblock Option Discovery for Autonomous Generation of Symbolic Knowledge.
\newblock In: International Conference of the Italian Association for Artificial Intelligence. Springer; 2022. p. 153--167.

\bibitem[\protect\citeauthoryear{Barto and Mahadevan}{2003}]{barto2003HRL}
Barto AG, Mahadevan S.
\newblock Recent advances in hierarchical reinforcement learning.
\newblock Discrete event dynamic systems. 2003;13(1):41--77.

\bibitem[\protect\citeauthoryear{Konidaris and Barto}{2009}]{Konidaris2009}
Konidaris G, Barto AG.
\newblock Skill discovery in continuous reinforcement learning domains using skill chaining.
\newblock In: Advances in neural information processing systems; 2009. p. 1015--1023.

\bibitem[\protect\citeauthoryear{Niel and Wiering}{2018}]{Niel2018}
Niel R, Wiering MA.
\newblock Hierarchical Reinforcement Learning for Playing a Dynamic Dungeon Crawler Game.
\newblock In: 2018 IEEE Symposium Series on Computational Intelligence (SSCI). IEEE; 2018. p. 1159--1166.

\bibitem[\protect\citeauthoryear{Rafati and Noelle}{2019}]{Rafati2019HRL}
Rafati J, Noelle DC.
\newblock Unsupervised Methods For Subgoal Discovery During Intrinsic Motivation in Model-Free Hierarchical Reinforcement Learning.
\newblock In: KEG@ AAAI; 2019. p. 17--25.

\bibitem[\protect\citeauthoryear{Parisi et~al.}{2021}]{Parisi2021}
Parisi S, Dean V, Pathak D, Gupta A.
\newblock Interesting object, curious agent: Learning task-agnostic exploration.
\newblock Advances in Neural Information Processing Systems. 2021;34:20516--20530.

\bibitem[\protect\citeauthoryear{Romero et~al.}{2021}]{Romero2021}
Romero A, Baldassarre G, Duro RJ, Santucci VG.
\newblock Analysing autonomous open-ended learning of skills with different interdependent subgoals in robots.
\newblock In: 2021 20th International Conference on Advanced Robotics (ICAR). IEEE; 2021. p. 646--651.

\bibitem[\protect\citeauthoryear{Bagaria et~al.}{2021}]{bagaria21a}
Bagaria A, Senthil JK, Konidaris G.
\newblock Skill Discovery for Exploration and Planning using Deep Skill Graphs.
\newblock In: Meila M, Zhang T, editors. Proceedings of the 38th International Conference on Machine Learning. vol. 139 of Proceedings of Machine Learning Research. PMLR; 2021. p. 521--531.
\newblock Available from: \url{https://proceedings.mlr.press/v139/bagaria21a.html}.

\bibitem[\protect\citeauthoryear{Veeriah et~al.}{2021}]{veeriah2021discovery}
Veeriah V, Zahavy T, Hessel M, Xu Z, Oh J, Kemaev I, et~al.
\newblock Discovery of options via meta-learned subgoals.
\newblock Advances in Neural Information Processing Systems. 2021;34:29861--29873.

\bibitem[\protect\citeauthoryear{Nau et~al.}{2004}]{NauGhallabTraverso2004}
Nau D, Ghallab M, Traverso P.
\newblock Automated Planning: Theory \& Practice.
\newblock San Francisco, CA, USA: Morgan Kaufmann Publishers Inc.; 2004.

\bibitem[\protect\citeauthoryear{Russell}{2010}]{russell2010artificial}
Russell SJ.
\newblock Artificial intelligence a modern approach.
\newblock Pearson Education, Inc.; 2010.

\bibitem[\protect\citeauthoryear{Chitnis et~al.}{2021}]{chitnis2021glib}
Chitnis R, Silver T, Tenenbaum JB, Kaelbling LP, Lozano-P{\'e}rez T.
\newblock Glib: Efficient exploration for relational model-based reinforcement learning via goal-literal babbling.
\newblock In: Proceedings of the AAAI Conference on Artificial Intelligence. vol.~35; 2021. p. 11782--11791.

\bibitem[\protect\citeauthoryear{Mikita et~al.}{2012}]{mikita2012interactive}
Mikita H, Azuma H, Kakiuchi Y, Okada K, Inaba M.
\newblock Interactive symbol generation of task planning for daily assistive robot.
\newblock In: 2012 12th IEEE-RAS International Conference on Humanoid Robots (Humanoids 2012). IEEE; 2012. p. 698--703.

\bibitem[\protect\citeauthoryear{Rodr{\'\i}guez-Lera et~al.}{2018}]{rodriguez2018generating}
Rodr{\'\i}guez-Lera FJ, Mart{\'\i}n-Rico F, Mateli{\'a}n-Olivera V.
\newblock Generating symbolic representation from sensor data: inferring knowledge in robotics competitions.
\newblock In: 2018 IEEE International Conference on Autonomous Robot Systems and Competitions (ICARSC). IEEE; 2018. p. 261--266.

\bibitem[\protect\citeauthoryear{Ganapini et~al.}{2022}]{frossi2022combining}
Ganapini MB, Campbell M, Fabiano F, Horesh L, Lenchner J, Loreggia A, et~al.
\newblock Combining Fast and Slow Thinking for Human-like and Efficient Navigation in Constrained Environments.
\newblock arXiv preprint arXiv:220107050. 2022;.

\bibitem[\protect\citeauthoryear{Garnelo et~al.}{2016}]{garnelo2016towards}
Garnelo M, Arulkumaran K, Shanahan M.
\newblock Towards deep symbolic reinforcement learning.
\newblock arXiv preprint arXiv:160905518. 2016;.

\bibitem[\protect\citeauthoryear{Garnelo and Shanahan}{2019}]{GARNELO201917}
Garnelo M, Shanahan M.
\newblock Reconciling deep learning with symbolic artificial intelligence: representing objects and relations.
\newblock Current Opinion in Behavioral Sciences. 2019;29:17 -- 23.
\newblock SI: 29: Artificial Intelligence (2019). \doi{https://doi.org/10.1016/j.cobeha.2018.12.010}.

\bibitem[\protect\citeauthoryear{Amado et~al.}{2018}]{amado2018goal}
Amado L, Pereira RF, Aires J, Magnaguagno M, Granada R, Meneguzzi F.
\newblock Goal recognition in latent space.
\newblock In: 2018 International Joint Conference on Neural Networks (IJCNN). IEEE; 2018. p. 1--8.

\bibitem[\protect\citeauthoryear{Asai and Fukunaga}{2018}]{Asai_Fukunaga_2018}
Asai M, Fukunaga A.
\newblock Classical Planning in Deep Latent Space: Bridging the Subsymbolic-Symbolic Boundary.
\newblock Proceedings of the AAAI Conference on Artificial Intelligence. 2018 Apr;32(1).
\newblock \doi{10.1609/aaai.v32i1.12077}.

\bibitem[\protect\citeauthoryear{Asai}{2019}]{asai2019unsupervised}
Asai M.
\newblock Unsupervised grounding of plannable first-order logic representation from images.
\newblock In: Proceedings of the International Conference on Automated Planning and Scheduling. vol.~29; 2019. p. 583--591.

\bibitem[\protect\citeauthoryear{Konidaris et~al.}{2018}]{konidaris2018skills}
Konidaris G, Kaelbling LP, Lozano-Perez T.
\newblock From skills to symbols: Learning symbolic representations for abstract high-level planning.
\newblock Journal of Artificial Intelligence Research. 2018;61:215--289.

\bibitem[\protect\citeauthoryear{Ghallab et~al.}{1998}]{Ghallab98}
Ghallab M, Howe A, Knoblock C, Mcdermott D, Ram A, Veloso M, et~al.: {PDDL---The Planning Domain Definition Language}.
\newblock Available from: \url{http://citeseerx.ist.psu.edu/viewdoc/summary?doi=10.1.1.37.212}.

\bibitem[\protect\citeauthoryear{Oddi et~al.}{2019}]{oddi2019learning}
Oddi A, Rasconi R, Cartoni E, Sartor G, Baldassarre G, Santucci VG.: Learning High-Level Planning Symbols from Intrinsically Motivated Experience.
\newblock ArXiv:1907.08313.

\bibitem[\protect\citeauthoryear{Sutton et~al.}{1999}]{SUTTON1999181_framework_b}
Sutton RS, Precup D, Singh S.
\newblock Between MDPs and semi-MDPs: A framework for temporal abstraction in reinforcement learning.
\newblock Artificial Intelligence. 1999;112(1):181--211.

\bibitem[\protect\citeauthoryear{Younes and Littman}{2004}]{younes2004ppddl1}
Younes HL, Littman ML.
\newblock PPDDL1. 0: An extension to PDDL for expressing planning domains with probabilistic effects.
\newblock Techn Rep CMU-CS-04-162. 2004;2:99.

\bibitem[\protect\citeauthoryear{Schmidhuber}{2010}]{schmidhuber2010formal}
Schmidhuber J.
\newblock Formal theory of creativity, fun, and intrinsic motivation (1990--2010).
\newblock IEEE transactions on autonomous mental development. 2010;2(3):230--247.

\bibitem[\protect\citeauthoryear{Santucci et~al.}{2010}]{Santucci2010biological}
Santucci VG, Baldassarre G, Mirolli M.
\newblock Biological Cumulative Learning through Intrinsic Motivations: A Simulated Robotic Study on the Development of Visually-Guided Reaching.
\newblock In: EpiRob. Citeseer; 2010. .

\bibitem[\protect\citeauthoryear{Barto et~al.}{2013}]{barto2013novelty}
Barto A, Mirolli M, Baldassarre G.
\newblock Novelty or surprise?
\newblock Frontiers in psychology. 2013;4:907.

\bibitem[\protect\citeauthoryear{Mirolli and Baldassarre}{2013}]{mirolli2013functions}
Mirolli M, Baldassarre G.
\newblock Functions and mechanisms of intrinsic motivations: The knowledge versus competence distinction.
\newblock Intrinsically motivated learning in natural and artificial systems. 2013;p. 49--72.

\bibitem[\protect\citeauthoryear{Oudeyer and Kaplan}{2009}]{oudeyer2009intrinsic}
Oudeyer PY, Kaplan F.
\newblock What is intrinsic motivation? A typology of computational approaches.
\newblock Frontiers in neurorobotics. 2009;p.~6.

\bibitem[\protect\citeauthoryear{Ester et~al.}{1996}]{10.5555/3001460.3001507}
Ester M, Kriegel HP, Sander J, Xu X.
\newblock A Density-Based Algorithm for Discovering Clusters in Large Spatial Databases with Noise.
\newblock In: Proceedings of the Second International Conference on Knowledge Discovery and Data Mining. KDD'96. AAAI Press; 1996. p. 226–231.

\bibitem[\protect\citeauthoryear{Cortes and Vapnik}{1995}]{10.1023/A:1022627411411}
Cortes C, Vapnik V.
\newblock Support-Vector Networks.
\newblock Mach Learn. 1995 Sep;20(3):273–297.

\bibitem[\protect\citeauthoryear{Rosenblatt}{1956}]{10.2307/2237390}
Rosenblatt M.
\newblock Remarks on Some Nonparametric Estimates of a Density Function.
\newblock The Annals of Mathematical Statistics. 1956;27(3):832--837.

\bibitem[\protect\citeauthoryear{Parzen}{1962}]{10.2307/2237880}
Parzen E.
\newblock On Estimation of a Probability Density Function and Mode.
\newblock The Annals of Mathematical Statistics. 1962;33(3):1065--1076.

\bibitem[\protect\citeauthoryear{Rolf et~al.}{2010}]{Rolf2010goalbabbling}
Rolf M, Steil JJ, Gienger M.
\newblock Goal babbling permits direct learning of inverse kinematics.
\newblock IEEE Transactions on Autonomous Mental Development. 2010;2(3):216--229.

\bibitem[\protect\citeauthoryear{Ecoffet et~al.}{2019}]{Ecoffet2019goexplore}
Ecoffet A, Huizinga J, Lehman J, Stanley KO, Clune J.
\newblock Go-explore: a new approach for hard-exploration problems.
\newblock arXiv preprint arXiv:190110995. 2019;.

\bibitem[\protect\citeauthoryear{Bharadhwaj et~al.}{2021}]{Bharadhwaj2021leaf}
Bharadhwaj H, Garg A, Shkurti F.
\newblock Leaf: Latent exploration along the frontier.
\newblock In: 2021 IEEE International Conference on Robotics and Automation (ICRA). IEEE; 2021. p. 677--684.

\bibitem[\protect\citeauthoryear{James et~al.}{2021}]{james2021autonomous}
James S, Rosman B, Konidaris G.
\newblock Autonomous learning of object-centric abstractions for high-level planning.
\newblock In: International Conference on Learning Representations; 2021. .

\end{thebibliography}

\end{document}